\documentclass[postprint,12pt]{elsarticle}%

\usepackage{etoolbox}
\usepackage{natbib}
\usepackage[utf8]{inputenc}
\usepackage{graphicx}
\usepackage{amsmath}
\usepackage{comment}
\usepackage{amsfonts}
\usepackage{amssymb}
\usepackage{subfigure}
\usepackage{multirow}
\usepackage{array}
\usepackage{subfigure}
\usepackage{booktabs}
\usepackage{cancel}
\usepackage[pdftex]{hyperref}
\usepackage{amsthm}
\usepackage{multicol}
\usepackage{enumerate} 
\usepackage{algorithm}
\usepackage{algpseudocode}
\usepackage{color}
\hypersetup{colorlinks=true,allcolors=black} 
\usepackage[dvipsnames,svgnames,x11names,hyperref,table]{xcolor}
\usepackage{epsfig}
\usepackage{array}
\usepackage{colortbl}
\usepackage{float}
\usepackage{cleveref}[2012/02/15]% v0.18.4; 
\crefformat{footnote}{#2\footnotemark[#1]#3}
\usepackage{tablefootnote}
\newcommand{\best}{Best}
\newcommand{\bem}{BEM}
\newcommand{\inv}{IEW}
\newcommand{\gem}{GEM}
\newcommand{\car}{Caruana}
\newcommand{\gs}{GS}
\newcommand{\rs}{RS}
\newcommand{\bo}{BO}
\newcommand{\pso}{PSO}
\newcommand{\hb}{HB}
\newcommand{\ridge}{Ridge}
\newcommand{\svr}{SVR}
\newcommand{\rfr}{RFR}
\newcommand{\randomforest}{RFR}
\newcommand{\ols}{OLS}
\newcommand{\fsr}{FSR}
\newcommand{\pcr}{PCR}
\newcommand{\pls}{PLS}
\newcommand{\bst}{BOOST}
\newcommand{\rbst}{RBOOST}
\newcommand{\aic}{AIC}
\newcommand{\aicc}{AICc}
\newcommand{\bic}{BIC}
\newcommand{\hqic}{HQIC}
\newcommand{\gmdl}{gMDL}
\newcommand{\icm}{ICM}
\newcommand{\total}{All}

\makeatletter
\def\ps@pprintTitle{%
 \let\@oddhead\@empty
 \let\@evenhead\@empty
 \def\@oddfoot{\footnotesize\itshape
       Postprint submitted to \ifx\@journal\@empty Elsevier
       \else\@journal\fi\hfill}
 \let\@evenfoot\@oddfoot}
\makeatother

\journal{Neurocomputing}

\begin{document}

\begin{frontmatter}

\title{Regularized boosting with an increasing coefficient magnitude stop criterion as meta-learner in hyperparameter optimization stacking ensemble}
	
\author[arcelor]{Laura Fdez-Díaz}
\ead{laura.fernandezdiaz@arcelormittal.com}
\author[uniovi]{Jos\'e Ram\'on Quevedo}
\ead{quevedo@uniovi.es}
\author[uniovi]{Elena Monta\~n\'es \corref{cor1}}
\ead{montaneselena@uniovi.es}

\cortext[cor1]{Corresponding author}

\address[arcelor]{ArcelorMittal (Spain)}
\address[uniovi]{Artificial Intelligence Center. University of Oviedo at Gij\'on, 33204 Asturias,  Spain \texttt{http://www.aic.uniovi.es}}

\begin{abstract}
Hyperparameter Optimization (HPO) aims to tune hyperparameters for a system in order to improve the predictive performance. Typically, only the hyperparameter configuration with the best performance is chosen after performing several trials. However, some works try to take advantage of the effort made when training all the models with every hyperparameter configuration trial and, instead of discarding all but one, they propose performing an ensemble of all the models. However, this ensemble consists of simply averaging the model predictions or weighting the models by a certain probability. Recently, some of the so-called Automated Machine Learning (AutoML) frameworks have included other more sophisticated ensemble strategies, such as the Caruana method or the stacking strategy. On the one hand, the Caruana method has been shown to perform well in HPO ensemble, since it is not affected by the issues caused by multicollinearity, which is prevalent in HPO. It just computes the average over a subset of predictions, previously chosen through a forward stepwise selection with replacement. But it does not benefit from the generalization power of a learning process. On the other hand, stacking approaches include a learning procedure since a meta-learner is required to perform the ensemble. Yet, one hardly finds advice about which meta-learner can be adequate. Besides, some possible meta-learners may suffer from problems caused by multicollinearity or need to be tuned in order to mitigate or reduce this obstacle. In an attempt to reduce this lack of advice, this paper exhaustively explores  possible meta-learners for stacking ensemble in HPO, free of hyperparameter tuning and able to mitigate the problems derived from multicollinearity as well as taking advantage of the generalization power that a learning process may include in the ensemble. Particularly, the boosting strategy shows promise in this context as a stacking meta-learner, since it satisfies the required conditions. In addition, boosting is even able to completely remove the effects of multicollinearity. This paper provides advice on how to use boosting as a meta-learner in the stacking ensemble. In any case, its main contribution is to propose an implicit regularization in the classical boosting algorithm and a novel non-parametric stop criterion suitable only for boosting and specifically designed for the HPO context. The existing synergy between these two improvements performed over boosting exhibits competitive and promising predictive power performance as a stacking meta-learner in HPO compared to other existing meta-learners and ensemble approaches for HPO other than the stacking ensemble. 

\end{abstract}
\end{frontmatter}
\section{Introduction} \label{sec:intro}
Hyperparameter Optimization (HPO) \cite{yu2020hyper} research rises from the need to find promising hyperparameter configurations in machine learning systems in order to achieve high predictive performance \cite{yang2020hyperparameter}. The hyperparameters determine the structure of the model and how the learning process will take place. They must be tuned before the learning process starts and they must be differentiate from the model parameters. Model parameters are estimated during the learning process, they configure the model itself and must be taken into account for making predictions.

HPO aims to obtain an optimal model that minimizes a prefixed loss function or maximizes a performance measure. Typically, the hyperparameter tuning process involves i) defining a model structure, ii) establishing the hyperparameters to be tuned and the domains for their values, iii) designing a hyperparameter value sampling method, iv) establishing an estimation procedure given an evaluation metric and, finally, v) configuring the final model. Among all these steps in hyperparameter tuning, researchers have focused their attention on developing promising strategies for hyperparameter sampling (step iii)) \cite{yu2020hyper, yang2020hyperparameter}. The common procedure in the literature for steps i), ii) and iv) involves respectively: checking several model structures, several hyperparameter domains and several evaluation procedures. Regarding the final configuration of the model (step v)), a typical approach would involve selecting the model with the best hyperparameter configuration based on the averaged evaluation metric estimations and then discarding the models with the rest of the hyperparameter configuration trials. Despite the selected model does provide the best estimation on average, it might not be  the best option overall. This means that more than one option might include at least some predictive power and may contribute to better overall performance. Hence, ensembling the models induced by the hyperparameter configuration trials \cite{mendes2012ensemble} seems like a straightforward favourable strategy to consider (one must not confuse this practice with using a sampling strategy in order to optimize an ensemble \cite{levesque2016bayesian}). In fact, there is a theoretical study based on ambiguity decomposition \cite{brown2005diversity} that shows that an adequate ensemble guarantees a better performance than the averaged performance of the individual models.

Some existing works have already explored the impact of performing ensemble in HPO \cite{escalante2010ensemble, momma2002pattern, alaa2018autoprognosis}, including the Basic Ensemble Method (\bem) \cite{perrone1992networks}, which computes the average of the model predictions, the Generalized Ensemble Method (\gem) \cite{perrone1992networks, shahosseini2022optimizing}, which performs regularized least squares regression under certain constraints,  or weighting by a probability as in the Inverse Expected Error Weighting (\inv) \cite{alaa2018autoprognosis}. Also, and more recently, some Automated Machine Learning (AutoML) systems include the option of performing an ensemble within their frameworks. However, only few of them do. In fact, AutoML systems mainly focus on other parts of the process, such as parallelizing or distributing the process, or on improving the performance of exploring encouraging hyperparameter configuration trials, or even on multi-output HPO \cite{survey2021multiobjetiveHPO}. The few systems that include ensemble in their scenarios do not perform an exhaustive study on them; instead they just contemplate the option whether to perform ensemble or not. These few AutoML systems either use the Caruana method \cite{caruana2004ensemble} for the ensemble (as in Auto-Gluon \cite{erickson2020autogluon}, Auto-Sklearn \cite{feurer2015efficient, feurer2022sklearn} or Auto-Pytorch \cite{zimmerPytorch2021}) or adopt stacking ensemble \cite{vanderlaan2007superlearner} (as in Auto-Weka \cite{kotthoff2017auto, kotthoff2019auto} or H2O \cite{pandey2019deep,ledell2020h2o}). Despite the Caruana method having shown to perform successfully due to the forward selection with replacement strategy, it only computes the simple average afterwards and does not profit from the generalization power of including a learning procedure. Unlike the Caruana method, the stacking procedure learns the ensemble, but in order for the system to learn the ensemble the meta-learner, must be carefully chosen. In fact, and to the best of our knowledge, the literature provides no guidelines as to which systems may be adequate in HPO. Furthermore, a quite recent survey argues this shortage of studies about stacking ensemble  \cite{kumar2022acomprehensive} for general purpose ensemble, hence, all the more reason for HPO.  Besides, if the system has hyperparameters, they must be carefully tuned in order to get good predictive performance, typically having to use an HPO procedure in turn. 

Bagging and boosting \cite{buhlmann2012bagging} are other general-purpose ensemble strategies widely used in the literature. The models to ensemble under these strategies are dynamically generated, and hence not applicable in the context of HPO, where the models to ensemble are defined beforehand and induced by the hyperparameter configuration trials.  Evolutionary algorithms \cite{pedrajas2005CIXL2} are another kind of general-purpose ensemble strategies, but they are usually overfilled with hyperparameters that need to be tuned. Other ensemble methods exist \cite{ren2016ensemble}, but they are developed specifically for certain cases, such as time series, neural networks, deep learning or multiple kernel learning \cite{ren2016ensemble}. Therefore, we discard these ensemble strategies, since they can not be adapted to HPO.

As a result, we shall now focus on stacking ensemble because (and unlike \bem, \inv\ and \car) this method takes advantage of the generalization power of ensemble through a learning process, despite a lack of guidelines about adequate non-hyperparametric meta-learners \cite{kumar2022acomprehensive}. As a matter of fact, one of the contributions of this paper is to review possible and adequate candidates as meta-learners for stacking ensemble in HPO; but before we delve into that, let us review the peculiarities of the HPO stacking ensemble context. The first peculiarity is that some of these hyperparameter trials may lead to excessively general models (underfitted models) that turn out highly similar predictions for all instances. The second peculiarity is that the values to ensemble may be excessively similar for some hyperparameter configurations. This happens because in the particular case of HPO the models to ensemble are all learned using the same machine learning system and, therefore, some variations in the values of the hyperparameters may not induce enough different models. These peculiarities give rise to the multicollinearity problem \cite{allen1997problem}. Muticollinearity emerges in a multivariable regression when the variables in the regression are highly correlated. This situation affects the accuracy in estimating the regression coefficients, producing skewed, misleading and unreliable results \cite{kiers2007comparison}. It commonly leads to overfitted models, hence, reducing the statistical power of the regression. In HPO, this problem gets particularly worse. The reason being, as stated before, the models to ensemble may be quite similar, and hence will provide similar predictions, which are in turn the values that must be ensembled. 

Ordinal Least Squares (\ols) \cite{abdi2003least} is an option as stacking meta-learner with no need to be tuned, but the above-mentioned peculiarities of the HPO ensemble context may cause overfitting. In fact, \ols\ is known to be highly affected by multicollinearity \cite{breiman1996stacked}, since it takes all the features to perform the regression. Adding constraints to \ols\ as a way of introducing a certain regularization procedure leads to \gem, which is slightly able to reduce the effects of multicollinearity with regard to the original \ols\, but not to a satisfactory degree. In fact, this approach has recently been proposed for ensembling in HPO \cite{shahosseini2022optimizing}. Some alternatives to \ols\ are Forward Stepwise Regression (\fsr) \cite{efroymson1960multiple}, Principal Component Regression (\pcr) \cite{merz98aprin}, Partial Least Squares (\pls) \cite{handbookPLS10vinzi} and Boosting (\bst) \cite{buhlmann2007boosting}. All of these approaches are iterative procedures that require predefining a number of iterations beforehand. However, this hyperparameter can be substituted by a stop criterion and, if this stop criterion is non-hyperparametric, then, the ensemble can be considered as non-hyperparametric. Other methods are able to overcome the the peculiarities of the HPO ensemble context but at the cost of tuning real-valued regularization hyperparameters. This is the case, for instance, of methods such as \ridge\ \cite{hothorn06unbiased}, Support Vector Regression (\svr) \cite{drucker1996support} or Random Forests Regression (\rfr) \cite{breiman01random}. In view of that, the first contribution of this paper is to discuss and explore the impact of \fsr, \pcr, \pls\ and \bst\ as meta-learners for stacking ensemble in HPO, analyzing different non-hyperparametric stop criteria so that the meta-learners become non-hyperparametric.

The second and main contribution of the paper is to include two improvements  in \bst\ in an attempt to exploit the specific potential of this method as meta-learner for stacking ensemble in HPO. Particularly, \bst\ can be promising in this context, since, unlike \fsr, \pcr\ and \pls, it carries out several regressions with just one feature each time, rather than including several features in the regression, as \fsr, \pcr\ and \pls\ do. The set of those different one-feature regressions are combined afterwards. On the one hand, performing a regression with just one feature allows using \ols\ as the base-learner regressor, removing the problems caused by the existing multicollinearity. On the other hand, the robust way adopted by \bst\  to combine the one-feature regressions makes it possible to detect collinear features as redundant, which enables significantly reducing, or even removing, the influence of these features in the ensemble.  However, this reduction may not be as promising as expected. This issue prompts one of the improvements to \bst, since one of the main contributions of this paper consists of including an implicit regularization in \bst\  in order to balance the influence of the collinear features in the ensemble. This practice leads to a method that we will call Regularized \bst\ (\rbst). The second improvement stemming from the main contribution of this paper consists of designing a novel stop criterion, which we will call Increasing Coefficient Magnitude (\icm) and which is specifically designed for \bst, taking advantage of its property about performing several regressions over just one feature. The result is that both novel improvements for \bst, namely, implicit regularization (\rbst) and the novel stop criterion (\icm), exert a synergy showing competitive and promising predictive power performance as meta-learner for stacking ensemble in HPO, compared to other existing non-hyperparametric meta-learners and other ensemble strategies different from stacking ensemble.

The rest of the paper is organized as follows: Section \ref{sec:related} describes some related work concerning the main existing AutoML frameworks and the main sampling strategies. Section \ref{sec:ensemble} deals with the ensemble paradigm. First,  several ensemble approaches of the literature are discussed more in depth. Then, it focuses on stacking ensemble and discusses non-hyperparametric meta-learners for it, along with several possible non-hyperparametric stop criteria. \bst\ as meta-learner in stacking ensemble is deeply detailed in Section \ref{sec:boosting}. In particular, this section also details the novel stop criteria \icm\ and  \rbst\ method as the result of carefully including an implicit regularization to \bst\  that was specifically designed for HPO. Experimental settings together with the description of the multicollinearity analysis carried out are described in Section \ref{sec:settingsmulticol}. Additionally, Section \ref{sec:results} presents the results and discusses the performance of the approaches. Finally, Section \ref{sec:conclusions} draws some conclusions and proposes some lines of research for future work.

\begin{table}[p]
\centering
\caption{Summary of sampling and ensemble strategies for the main AutoML and CASH frameworks\label{tab:frameworks}}
\scalebox{0.9}{
\begin{tabular}{llll}
\hline
&\textbf{Framework}&\textbf{Sampling}&\textbf{Ensemble}\\ 
%&&\textbf{strategy}&\textbf{strategy}\\ 
\hline
\multirow{18}{*}{AutoML}&ATM \cite{swearingen2017atm} &BO, multi-armed bandit&-\\
&Auto-Gluon \cite{erickson2020autogluon}&-&Caruana \cite{caruana2004ensemble}\\ 
&Auto-Pytorch \cite{zimmerPytorch2021} &BO, HB&Caruana \cite{caruana2004ensemble}\\
&Auto-Sklearn \cite{feurer2015efficient, feurer2022sklearn} &BO, Successive Halving&Caruana \cite{caruana2004ensemble}\\ 
&Auto-Weka \cite{kotthoff2017auto, kotthoff2019auto} &BO&Stacking \cite{vanderlaan2007superlearner}\\ 
&Hyperopt-Sklearn \cite{komer2014hyperopt, bergstra2015hyperopt} &BO&-\\ 
&H2O \cite{pandey2019deep, ledell2020h2o}  &GS, RS &Stacking \cite{vanderlaan2007superlearner}\\ 
&TPOT \cite{olson2019tpot} &Genetic Algorithm&-\\
&TPOT-NN \cite{romano2021tpot} &Genetic Algorithm&-\\
&MANGO \cite{sandha2020mango} &BO&-\\
&Syne-Tune \cite{salinas2022syne} &BO, HB, Population-based&-\\
&Hyper-Tune \cite{yang2022hypertune} &Improved BO& -\\
&Google Vizier \cite{golovin2017google} &BGPB\tablefootnote{Batched Gaussian Process Bandits}, others&-\\
&\multirow{2}{*}{Ray-Tune \cite{liaw2018raytune}} &GS, RS, BO, HB&\multirow{2}{*}{-}\\
&&Blend-search, BO Dragonfly&\\
&OpenBox \cite{yang2021openbox}&BO, PRF\tablefootnote{Probabilistic Random Forest}&-\\ 
&ASHA \cite{li2018asystem}&Succesive Halving&-\\ 
&MFest-HB \cite{yang2021mfeshb}&BO, HB&-\\
\hline
&\textbf{Framework}&\textbf{Sampling}&\textbf{Ensemble}\\ 
\hline
\multirow{7}{*}{CASH}&Optunity \cite{claesen2014easy}&\pso&-\\
&\multirow{2}{*}{BOHB  \cite{falkner2018bohb}}& Bayesian&\multirow{2}{*}{-}\\
&&Hyperband\\
&SMAC \cite{van2018hyperparameter} &Bayesian&-\\
&RoBo \cite{klein2017robo}& Bayesian&-\\
&\multirow{2}{*}{BTB  \cite{gustafson2018bayesian}}&Bayesian&\multirow{2}{*}{-}\\
&&Multi-armed Bandit\\
\hline
\end{tabular}}
\end{table}

\section{Related Work} \label{sec:related}
A key step in HPO is hyperparameter sampling. Great efforts have been made in the literature to design promising sampling strategies, becoming the main aspect in the HPO field \cite{yu2020hyper, yang2020hyperparameter}. Some hyperparameter sampling strategies do not take into account model evaluations to obtain different hyperparameter configuration samples; instead, each sample is drawn independently of the rest. This property makes it possible to learn several configuration trials in parallel but incurs the risk of wasting time on exploring poorly performing configurations. This is the case of  Grid Search (\gs) \cite{claesen2015hyperparameter, sambridge1986novel} and Random Search (\rs) \cite{bergstra2012random}. \gs\ \cite{claesen2015hyperparameter, sambridge1986novel} is a simple approach that takes a finite set of values for each hyperparameter and computes the Cartesian product of them to configure a grid of trials to be checked. \rs\ \cite{bergstra2012random} randomly draws a predefined number of trials according to certain distribution following a Monte Carlo technique. Unlike those methods that sample each configuration trial independently, others include a guided search involving a model evaluation of the current sample in order to draw the next, as in Bayesian Optimization (\bo) \cite{snoek2012practical}, Particle Swarm Optimization (\pso) \cite{shi1998parameter} and Hyperband (\hb) \cite{li2017hyperband}. \bo\ \cite{snoek2012practical}  is a well-known and successful optimization approach \cite{jones1998efficient} that performs a balance between exploration (taking other hyperparameter values) and exploitation (taking information from the hyperparameters already explored) in order to avoid falling into a local minimum.  \pso\ \cite{shi1998parameter} is a population-based method that simulates a biological behaviour among particles that has also been successfully applied in HPO \cite{lorenzo2017particle}. The particles in \pso\ just cooperate rather than mutate or crossover. This provides information to guide the search, but it must be properly initialized to minimize the risk of leading to a local rather than to a global optimum. \hb\ \cite{li2017hyperband} is a bandit-based technique that improves on the successive halving method. It does so by dynamically choosing hyperparameter configurations in an attempt to establish a trade-off between the number of configurations and the available resources (such as time). This way, half of the poorly performing configurations are eliminated each time, while the other half are kept.

Nowadays, HPO is one of the core parts of the AutoML frameworks \cite{zoller2021benchmark}. While HPO tries only to provide a predictive model by optimizing their hyperparameters, AutoML goes further and does something more than this. Particularly, AutoML covers solving all the tasks a researcher must tackle ,obtaining a final solution from the data, trying to avoid requiring expertise assistance. Even though this task includes pre-processing, feature selection and extraction before the predictive model is induced it also calls for interpretability and decision making after the predictive model is induced. Combined Algorithms Selection and Hyperparameter Optimization (CASH) \cite{wangautomodel2020} is currently a top field of research that goes a step beyond the task of HPO as well, since it also selects a suitable system that provides the model in addition to the hyperparameters. However, CASH environments do not automate as many tasks as an AutoML framework.  In fact, it is quite common for AutoML frameworks to be built over a CASH environment. 

Some AutoML frameworks have been proposed in the literature. For instance, Auto-Weka \cite{kotthoff2017auto, kotthoff2019auto} is an AutoML environment built on top of WEKA models. Auto Tuned Models (ATM) \cite{swearingen2017atm}, Auto-Sklearn \cite{feurer2015efficient, feurer2022sklearn} and Tree-based Pipeline Optimization Tool (TPOT) \cite{olson2019tpot} are frameworks that use the scikit-learn library \cite{pedregosa2011scikit}. Hyperopt-sklearn \cite{komer2014hyperopt, bergstra2015hyperopt} is based on Auto-Weka applied to scikit-learn. Some frameworks focus specifically on neural networks such as Auto-Pytorch \cite{zimmerPytorch2021} and TPOT-NN  \cite{romano2021tpot} (a particular version of TPOT). Auto-Gluon \cite{erickson2020autogluon} successfully includes a multilayer combination of models for image, text, time series, and tabular data. H2O \cite{pandey2019deep, ledell2020h2o} is an open source, in memory, distributed, fast and scalable commercial platform also suitable to be managed by non-experts in machine learning.  Another recent AutoML framework is MANGO \cite{sandha2020mango}, which is an open-source Python library able to parallelize HPO on a distributed cluster. Syne-Tune \cite{salinas2022syne} is an open-source Python library as well, but for large-scale distributed hyperparameter and neural architecture optimization. Also, Ray-Tune \cite{liaw2018raytune} is specifically designed for distributed model selection. Additionally, ASHA \cite{li2018asystem} proposes an asynchronous successive halving algorithm in order to improve the efficiency for numerous parallel evaluations. Even more recently, Hyper-Tune \cite{yang2022hypertune} has included improvements in regard to optimizing the BO, such as automatic resource allocation, asynchronous scheduling and multi-fidelity optimizer. MFest-HB \cite{yang2021mfeshb} proposes a new sampling strategy including multi-fidelity learning to HB sampling strategy, which improves the Bayesian Optimization and Hyperband (BOHB) \cite{falkner2018bohb}. Finally, other systems are Google Vizier \cite{golovin2017google} and OpenBox \cite{yang2021openbox}, which both include transfer learning and early stopping to improve the hyperparameter search, but Google Vizier only supports traditional black-box optimizations, whereas OpenBox can cope with multiple objectives and constraints.  The fact is that all of these AutoML systems focus on improving the configuration trials generation or on parallelizing or distributing the computations, and only some of them include ensemble after learning the models with the different  generated configuration trials. Particularly, only Auto-Gluon, Auto-Pytorch and Auto-Sklearn include the Caruana method for ensemble, while only Auto-Weka and H2O allow the possibility of performing stacking, but without any advice about which meta-learner is adequate to use. Concerning CASH environments, Auto-Sklearn \cite{feurer2015efficient, feurer2022sklearn} and Hyperopt-sklearn \cite{komer2014hyperopt, bergstra2015hyperopt} also deal with CASH for supervised machine learning. Optunity \cite{claesen2014easy}, Bayesian Optimization and Hyperband (BOHB) \cite{falkner2018bohb}, Sequential Model-based Algorithm Configuration (SMAC) \cite{van2018hyperparameter}, Robust Bayesian Optimization (RoBo) \cite{klein2017robo}, Bayesian Tuning and Bandits (BTB) \cite{gustafson2018bayesian} are other popular CASH frameworks. Table \ref{tab:frameworks} summarizes the sampling and ensemble strategies supported by all these AutoML and CASH environments. 

\section{Ensemble in Hyperparameter Optimization}\label{sec:ensemble}
A general definition of ensemble learning that covers supervised (classification and regression) and unsupervised learning can be the process of integrating a set of models in order to provide a final prediction \cite{mendes2012ensemble, ren2016ensemble}. Formally, this integration process $\mathcal{I}$ can be defined for a given instance $x$ as
\[
f (x)=\mathcal{I}(f_1 (x),....,f_p (x))
\]
where $p$ is the number of models to ensemble, $\{f_i (x)\}_{i=1}^p$ is the set of models to ensemble and $f (x)$ is the function obtained after the integration process.

The integration process is commonly assumed to involve a linear combination or fusion of the individual models. Hence, $f (x)$ can be rewritten as
\[
f(x)=\sum_{i=1}^p h_i(x)\cdot f_i(x) 
\]
where $\{h_i (x)\}_{i=1}^p$ is a set of functions that grant weights to the individual models  $\{f_i (x)\}_{i=1}^p$.

Diversity is a key issue in ensemble learning \cite{tang2006analysis}. However, among the existing kinds of diversity, hyperparameter diversity is the one that fits HPO, since in HPO different hyperparameter configuration trials are the ones that provide the different models to ensemble \cite{ren2016ensemble}.
 
The common practice establishes the weighting functions $\{h_i (x)\}_{i=1}^p$ as constants, that is, $\{h_i (x)\}_{i=1}^p=\{\alpha_i\}_{i=1}^p$. Non-constant weighting functions have also been studied. Static methods (methods that define non-constant weighting functions during learning) either split the input space by assigning models to predefined regions \cite{peter2012bagging} or perform static selection defining areas of expertise for the models \cite{kuncheva2002switching}. dynamic methods (methods that define non-constant weighting functions in prediction time), on the other hand, do it by searching for similar instances in the training set, typically via k-nearest neighbour approaches \cite{rooney2004dynamic}. In any case, both types require hyperparameter tuning, for subsampling in the case of the static methods and for the k-nearest neighbour based approaches in the case of dynamic methods. 

Hence, this paper will focus solely on constant-weighting functions, and only on those that do not require adjusting hyperparameters within them, since one of the goals of this paper is precisely to propose a non-hyperparametric ensemble procedure. Section \ref{sec:ensemblestrategies} reviews existing ensemble strategies, including stacking. Section \ref{sec:metalearners} discusses meta-learners for stacking ensemble, for which a stop criterion must be stated. Finally, Section \ref{sec:stops} makes a review and discussion of existing stop criteria for meta-learners in stacking ensemble.

\subsection{Review of existing ensemble strategies in Hyperparameter Optimization}\label{sec:ensemblestrategies}
\bem\ \cite{perrone1992networks} has been employed for ensemble in HPO and provides constant-weighting functions without tuning hyperparameters, just by computing the simple average of the individual predictions $\{f_i(x)\}_{i=1}^p$. Hence, the $\{\alpha_i\}_{i=1}^p$ are all equal to the constant $1/p$ for all $i=1,\dots,p$. The ensemble function is then $f(x)=\sum_{i=1}^p \frac{1}{p}\cdot f_i(x)=\frac{1}{p}\cdot \sum_{i=1}^p  f_i(x)$. The \inv\ strategy \cite{wang2003mining}, which consists of establishing the weights $\{\alpha_i\}_{i=1}^p$ as inversely proportional to the expected error of $\{f_i (x)\}_{i=1}^p$,  has also been employed as ensemble method in HPO and also provides constant-weighting functions free of hyperparameter tuning. The Caruana strategy \cite{caruana2004ensemble} is an appealing approach that goes along the same line as the previous methods, and has shown promising results recently and has been included in some of the few AutoML systems that provide ensemble in their frameworks \cite{erickson2020autogluon, feurer2015efficient, zimmerPytorch2021}. It differs from the above-mentioned methods in that it performs an ensemble selection first, or, in other words, it establishes some weights to be zero beforehand. More in detail, first the best models that will not be weighted by zero are selected with replacement and the simple average is then computed. As a result, the weights of each model depend on the number of times the model was selected.

These strategies do not require hyperparameter tuning. Futhermore, multicollinearity does not affect them. In the case of \bem, this is so because the weights are constant and chosen independently of the prediction values. In the case of \inv, the weight for each prediction only depends on the prediction of this model, hence, it is chosen independently of the rest. Finally, in the  Caruana strategy zero weight is implicitly assigned to some predictions, namely, to those that are not involved in the selection under a replacement procedure. However, these ensemble methods, do not include a learning process that may add generalization power to the ensemble. Bagging, boosting and stacking \cite{buhlmann2012bagging}  are typical ensemble strategies that include a learning procedure in the process, have been widely used for many applications \cite{zhang2022areview} and have been recently stated as the most promising kind of ensemble approaches regarding data, algorithm or output level manipulation approaches \cite{kumar2022acomprehensive}. Among the three strategies, stacking is the only one suitable to be applied in HPO and has been included in some of the few AutoML systems that use ensemble in their frameworks. Bagging and boosting are not suitable for ensemble in HPO, since models for these approaches are dynamically generated, whereas the models in HPO are learned beforehand according to the range of hyperparameter configuration trials. Apart from the conventional methods for ensemble (bagging, boosting and stacking), other ensemble methods specifically designed for certain situations are available \cite{ren2016ensemble}. This is the case of decomposition based methods, typically adopted for time series datasets, which can be classified into divide-and-conquer and hierarchical ensemble methods. The main concept is to decompose the time series into a  collection of time series motivated by its seasonal properties. Hence, these methods are not applicable to general-purpose datasets. There are also multi-output optimization ensemble methods, which try to optimize several performance measures and typically adopt evolutionary algorithms to find the Pareto front of the individual models. We discard these methods, since optimizing several performance measure falls out of the scope of this work. Besides, evolutionary algorithms have more than plenty of hyperparameters to tune. Negative correlation ensemble method has been specifically designed for neural networks, where all the individual models are trained simultaneously using penalty terms in the respective error functions. This method is not applicable in our context, since the individual models are trained taking into account the configuration trials that the sampling strategies generate. Deep learning and multiple kernel learning based ensemble methods are also available, but they typically require tuning a considerable number of hyperparameters.

Let us now focus on stacking ensemble. Stacking \cite{vanderlaan2007superlearner}, also called stacked ensemble, stacked regression or superlearning, aims to find an optimal combination of the models $\{f_i(x)\}_{i=1}^p$, while providing constant-weights, but including a learning process in the ensemble, typically known as a second-level meta-learner. The inclusion of this meta-learner provides the ensemble strategy with a promising generalization power. However, the main drawback is establishing an adequate meta-learner for HPO, since it may be affected by the existing multicollinearity. Currently, the literature does not provide advice on this issue. In fact, in a very recent survey of ensemble methods \cite{kumar2022acomprehensive}, it is stated that stacking has not been extensively studied so far and it is suggested as a future research line. Furthermore, this meta-learner may have hyperparameters to be tuned, as it is the case with \ridge\ \cite{hothorn06unbiased}, \svr\ \cite{drucker1996support} or \rfr\ \cite{breiman01random}, in order to avoid, or at least mitigate, the problems derived from the multicollinearity.  An alternative could consist of using classical \ols, which is highly affected by multicollinearity, or even, classical \ols\ with constraints such as the weights to be positive ($\{\alpha_i\ge 0\}_{i=1}^p$) and sum to one ($\sum_{i=1}^p\alpha_i=1$) in order to express the generalized error, which leads to the method called \gem\ \cite{perrone1992networks}. \gem\ also encounters multicollinearity, but it deals with it by imposing the constraints of the weights to be positive and sum to one. In fact, it is one of the methods that has been recently applied to ensemble in HPO \cite{shahosseini2022optimizing}.
Section \ref{sec:metalearners} discusses possible meta-learners for stacking ensemble. 

\subsection{Discussing possible meta-learners for stacking ensemble in Hyperparameter Optimization}\label{sec:metalearners}
As commented before, there is no advise in the literature about which meta-learners may be suited to ensemble stacking and, in particular, to HPO ensemble stacking. Therefore, some possibilities are exposed in this section. Apart from \ols\ \cite{abdi2003least} and \gem, which are highly affected by the multicollinearity problem, and discarding methods that require to tune hyperparameters, \fsr\ \cite{efroymson1960multiple}, \pcr\ \cite{merz98aprin}, \pls\ \cite{handbookPLS10vinzi} and \bst\ \cite{buhlmann2007boosting} remain possible meta-learners for stacking ensemble. At this point, we shall clearly state that \bst\ is only used here as a meta-learner for stacking ensemble, and not as an HPO ensemble \cite{buhlmann2012bagging}.

\fsr\  involves starting with no features in the model, testing the addition of one single feature at a time, using a chosen model fit criterion that adds the feature (if any) that contributes the most statistically significant improvement of the fit, and repeating this process until a stop criterion is satisfied. If every feature is included, then \fsr\ becomes \ols. \fsr\ has been recently adopted for ensemble in HPO \cite{wenzel2020hyperparameter}. However, this work does not focus on the ensemble itself; it just adopts \fsr\ for ensemble and takes the number of iterations as a stop criterion, hence, adding an additional hyperparameter. The work focuses on deep neural networks and on the claims noticeable benefit when combining (ensembling) different hyperparameter values (coming from a RS sampling strategy) together with different possible initializations of the deep neural network.

\pcr\ computes the so-called principal components, which are the eigenvectors of the covariate matrix, which in turn are the directions of the axes of the most variance, and hence, provide the most information. The principal components are uncorrelated and the information of the original features is expected to be squeezed or compressed into the first components. In this way, the first principal component accounts for the largest possible variance, and consequently, for the most information. The second principal component is uncorrelated with the first principal component and accounts for the next highest variance, and so on until a stop criterion is satisfied. Finally, a regression, typically using an \ols,  is performed using the first principal components as features instead of using the original features. Despite this strategy may seem to reduce the multicollinearity because the regression is performed over uncorrelated and transformed features, the main drawback is that no information about the target is taken, so the components are taken in an unsupervised way. Therefore, there is no guarantee on whether the principal components will be related to the target. In this sense, \pls\ and \pcr\ work similarly, but in \pls\ the principal components try to extract those features that explain as much as possible the covariance between the features and target, rather than the variance between the features. Then, unlike \pcr, \pls\ takes into account the relationship between the features and the target, making the principal components closer to the target. This property makes \pls\ a stronger meta-learner than \pcr. 

Finally, \bst\ works quite differently from \fsr, \pcr\ and \pls, since \bst\ performs a regression using just one feature each time, therefore, completely removing the multicollinearity problem. This is a promising property, not only because it makes \bst\ free of multicollinearity, more importantly it 
allows including regularization strategies and stop criteria that involve just one feature. Section \ref{sec:boosting} formally explains the \bst\ strategy. It also details our novel proposal \rbst\ as a meta-learner in HPO stacking ensemble, which consists of adding an implicit regularization in \bst\ (see Section \ref{sec:regul}). Finally, our novel stop criterion \icm\ is exhaustively and specifically built for \bst\ and \rbst\ (see Section \ref{sec:icm}).

\subsection{Discussing several possible stop criteria for the meta-learners for stacking ensemble in Hyperparameter Optimization}\label{sec:stops}
All \fsr, \pcr, \pls\ and \bst\ require a stop criterion as an alternative to the number of iterations (features), or, in the case of \pcr\ and \pls, the number of principal components. Several non-hyperparametric stop criteria are specifically available in the literature for \fsr\ \cite{banks17applied}, \pcr\ \cite{bai2018consistency} and \pls\  \cite{nengsih19determining}. The process works as follows. A stop criterion of this kind is computed for all possible values of the number of features (in the case of \fsr) or number of components (in the case of \pcr\ or \pls). The possible values in both cases will range from 1 to $p$, where $p$ is the number of models involved in the ensemble. Then, the best features from \fsr, (or components using \pcr\ or \pls) are obtained for each of these possible values of the number of features (or components). Next, an \ols\ is carried out over these features (or components), which yields a performance score. After that,  the stop criterion is computed from the number of features or components, the features or components themselves and the performance score. Finally,  the best option will be the number of features or components that provides the best stop criterion value. These stop criteria are Akaike Information Criterion (\aic) \cite{akaike1998information}, Akaike Information Criterion corrected (\aicc) \cite{hurvich1989regression}, Bayesian Information Criterion (\bic) \cite{schwartz1997stochastic}, Hannan-Quinn Information Criterion (\hqic) \cite{hannan1979determination} and generalized Minimum Description Length (\gmdl) \cite{hansen2001model}. \aic\ determines the relative information value of the model using the maximum likelihood estimation and the number of features. The best-fit model according to \aic\ is the one that explains the greatest amount of variation using the fewest possible features. \aicc\ introduces a correction into \aic\ in order to avoid overfitting when the number of instances is small in comparison with the number of features. \bic\ is similar to \aic, but it penalizes more aggressively the number of instances. \hqic\ introduces a correction over \bic\, to smooth the influence of the number of instances. Finally, \gmdl\ combines AIC and BIC and tries to adaptively select the best between the two. All these stop criteria are designed for general purpose regression rather than for HPO ensemble. Specifically, they penalize the number of features and tend to discard adding new features if the prediction performance hardly improves. However, one of the challenges of HPO ensemble is to include in the final model the maximum information contained in the models trained with the variety of hyperparameter configuration trials, even if the performance with fewer models may be accurate enough. In this sense, our proposed stop criterion \icm\ does not penalize the number of features and therefore allows including information coming from both the previous or new features until an overfitting situation is detected. Section \ref{sec:icm} exposes in detail how this novel stop criterion was deduced.

\begin{figure}[p]
\centering
	\includegraphics[trim={4cm 0 3cm 0}, scale=0.6]{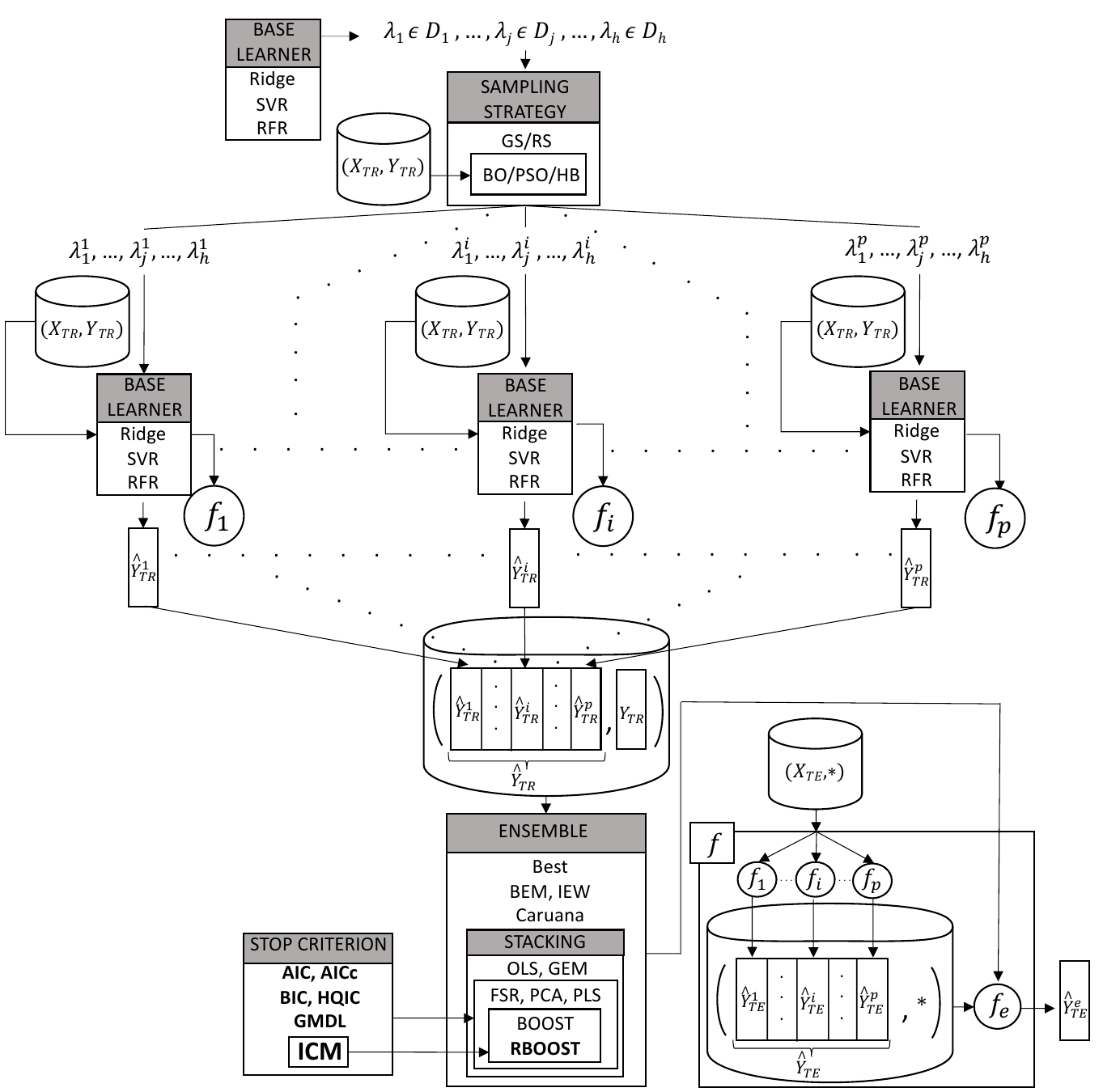}
%	\caption{Overall schema of the process. A sampling strategy generates several hyperparameter configurations for inducing several models using a base-learner. These models are aggregated using an ensemble method. In case of stacking ensemble, a meta-learner is required, some of them need a stop criterion}
		\caption{Overall scheme of the process. A sampling strategy (\gs, \rs, \bo, \pso\ or \hb) generates several hyperparameter configurations for inducing several models using a base-learner (\ridge, \svr\ or \rfr). These models are aggregated using an ensemble method (\best, \bem, \inv, \car\ or stacking). In the case of stacking ensemble, a meta-learner (\ols, \gem, \fsr, \pcr, \pls, \bst\ or \rbst) is required, and some require a stop criterion
		(\aic, \aicc,\bic, \hqic, \gmdl\ or \icm)}
	\label{fig:overallschema}
\end{figure}

\subsection{Overall process of ensemble in Hyperparameter Optimization}
This section summarizes the whole process of HPO with ensemble. Particularly, Figure \ref{fig:overallschema} illustrates the integration of the sampling strategies mentioned in Section \ref{sec:related} (\gs, \rs, \bo, \pso\ and \hb). These sampling strategies generate several configuration trials $\{{\lambda_1^i, \dots, \lambda_j^i,\dots, \lambda_h^i\}}_{i=1}^p$ for the hyperparameters $\lambda_1, \dots, \lambda_j,\dots, \lambda_h$ of certain base-learners (\ridge, \svr\ and \rfr). The base-learners are trained from an $(X_{TR}, Y_{TR})$ dataset taking into account these configuration trials $\{{\lambda_1^i, \dots, \lambda_j^i,\dots, \lambda_h^i\}}_{i=1}^p$, leading to the set of models $\{f_i\}_{i=1}^p$. Then, the models $\{f_i\}_{i=1}^p$ are applied to $X_{TR}$ (typically using a cross validation strategy) to provide the set of predictions $\hat{Y}_{TR}=\{\hat{Y}^i_{TR}\}_{i=1}^p$. Hence, the data set $(\hat{Y}_{TR}, Y_{TR})$ feeds an ensemble strategy outlined in Section \ref{sec:ensemblestrategies} (\bem, \inv, \car\ and staking ensemble). As far as \best\ is concerned, it applies the classical procedure in HPO of choosing the best model from $\{f_i\}_{i=1}^p$ according to a loss function. Focusing on staking ensemble, several possible meta-learners discussed in Section \ref{sec:metalearners} (\ols, \gem, \fsr, \pcr, \pls, \bst\ and our novel proposal \rbst, built on the basis of \bst) can be applied. Concerning \fsr, \pcr, \pls, \bst\ and even \rbst, a non-hyperparametric stop criterion must be established in order for the meta-learner to be non-hyperparametric. This stop criterion can be one of those presented in Section \ref{sec:stops} (\aic, \aicc, \bic, \hqic, \gmdl\ and of course our novel stop criterion \icm). Finally, an ensemble model $f_e$ is induced. As a result, the overall model $f$ is formed by the $\{f_i\}_{i=1}^p$ models induced by the different configuration trials and the ensemble model $f_e$. Hence a test dataset $X_{TE}$ is applied to the configuration trial models $\{f_i\}_{i=1}^p$, whose predictions are ensembled using the ensemble model $f_e$ providing the prediction $\hat{Y}_{TE}^e$ for $X_{TE}$.

\section{Boosting as meta-learner for stacking ensemble in HPO}\label{sec:boosting}
This section discusses \bst\ and the proposed implicit regularization for \bst, which leads to the \rbst\ method as a meta-learner in stacking ensemble (see Section \ref{sec:regul}). Additionally, this section exposes in detail how the novel stop criterion \icm, specifically designed for \bst\ and \rbst, is derived (see Section \ref{sec:icm}). In fact, it is not applicable to other meta-learners. 

We shall begin with detailing \bst. Let $\mathcal{F}=\{f^i (x)\}_{i=1}^p$ be the set of $p$ features that describe the predictions provided by the model induced from the different hyperparameter configuration trials. Initially, the set of features $\mathcal{S}$ for the ensemble is empty, that is, $\mathcal{S}^{(0)}=\emptyset$, since the algorithm follows a forward-search strategy. In each stage $j$, a set of regression procedures involving one single feature is performed: one regression per feature in $\mathcal{F}$ (a feature is selected with replacement). A feature $f^{i^{*,(j)}}(x)$ of $\mathcal{F}$ is selected in stage $j$ according to a certain criterion in terms of a loss function $\mathcal{L}$ and included in $\mathcal{S}$, that is, $\mathcal{S}^{(j)}=\mathcal{S}^{(j-1)}\cup \{f^{i^{*,(j)}} (x)\}$. The target for performing the set of regressions in each stage remains constant for the set of regressions performed in each stage, although it does vary from one stage to another. Hence, the bias is also corrected from one stage to another. In the first stage, the target for the set of regressions is the original one, that is, $r^{(0)}=y$. Then, in each stage, the target for the next stage $r^{(j)}$ is computed as the difference between current stage's target $r^{(j-1)}$ and the prediction performed using the regression model that was induced with the feature selected in the actual stage $h_{f^{i^{*,(j)}}(x)}(f^{i^{*,(j)}}(x))$. More in detail, the set of regressions performed in certain stage $j$ is 
\[
\{h_{f^{i}(x)}^{(j)}(f^{i}(x))=r^{(j-1)} : f^i (x)\in \mathcal{F}\}
\]
%\[
%\{\alpha_i^{(j)} \cdot f^i(x)+\beta_i^{(j)}=r^{(j-1)} : f^i (x)\in \mathcal{F}\}
%\]
%$\alpha_{i^{*,(j)}}^{(j)}\cdot f^{i^{*,(j)}} (x)+\beta_{i^{*,(j)}}^{(j)}$
where the initial residual is $r^{(0)}=y$ and the residual in stage $j$ is defined in terms of the selected feature $f^{i^{*,(j)}} (x)$ in stage $j$, that is, $r^{(j)}=r^{(j-1)}- h_{f^{i^{*,(j)}}(x)}(f^{i^{*,(j)}}(x))$. The process continues until the stop criterion is satisfied. Consequently, \bst\ builds a family of functions $\{g^{(j)}(f(x))\}_{j=1}^{\cdots}$ in a stage-wise rather than in a step-wise procedure, such as
\[
g^{(j)}(f(x))=g^{(j-1)}(f(x))+ h^{(j)}(f(x))
\]
where $g^{(0)}(f(x))=0$. 

Considering linear regression, the model induced in each stage takes the form 
\[
h_{f^{i^{*,(j)}}(x)}(f^{i^{*,(j)}}(x))=\alpha_{i^{*,(j)}}^{(j)}\cdot f^{i^{*,(j)}} (x)+\beta_{i^{*,(j)}}^{(j)}
\]
where $\alpha_{i^{*,(j)}}^{(j)}$ and $\beta_{i^{*,(j)}}^{(j)}$ are the regression coefficients for the $f^{i^{*,(j)}} (x)$ feature taken in stage $j$. The features in $\mathcal{S}$ after the process ends will be the features taken in the ensemble with the weights $\alpha_{i^{*,(j)}}^{(j)}$ successively computed in the process. As commented before, the process selects adequate features in each stage with replacement, which means that a selected feature in a certain stage might be chosen again in successive stages. The respective weights $\alpha_{i^{*,(j)}}^{(j)}$ for this kind of features are accumulated to provide a unique weight to the feature. 
\[
\alpha_{f^{i^*}(x)}=\sum_{
\begin{array}{c}
f^{i^{*,(j)}}(x)\in \mathcal{S}\\
f^{i^{*,(j)}}(x)=f^{i^{*}}(x)
\end{array}
}\alpha_{i^{*,(j)}}^{(j)}
\]
Bias is included in the procedure since it is not possible to ensure that the features are meaningfully unbiased. All the bias $\beta_{i^{*,(j)}}^{(j)}$ of the regression in each stage are also successively accumulated. Then,
\[
\beta=\sum_{f^{i^{*,(j)}}(x)\in \mathcal{S}}\beta_{i^{*,(j)}}^{(j)}
\]
Therefore, including the bias in the procedure implicitly alters the expression of the ensemble when compared to the one displayed in the previous section. The new expression for the ensemble will be
\[
f(x)=\sum_{f^{i^{*}} (x)\in \mathcal{S^*}} \alpha_{f^{i^{*}}}\cdot f^{i^{*}}(x) + \beta 
\]
where $\mathcal{S^*}$ is the set of selected features, where $f^{i^{*}} (x)$ represents a different feature (without replacement).

\begin{figure}[p]
      \centering
\begin{algorithm}[H]\small
\caption{\bst\ for stacking ensemble}
\label{alg:bst}
\renewcommand{\algorithmicrequire}{\textbf{Input:}}
\renewcommand{\algorithmicensure}{\textbf{Output:}}
\newcommand{\Break}{\State \textbf{break} }
\begin{algorithmic}[1]
\Function{boosting}{}
\begin{flushleft}
       \textbf{Input:} $\{f^{i} (x)\}_{i=1}^p$ features of the ensemble (model predictions), $y$ the original target$,\mathcal{L}$ a loss function\\
        \textbf{Output:} $\mathcal{S}$, $\mathcal{A}$, $\beta$
\end{flushleft}
\State $\mathcal{A} \leftarrow  \emptyset$ 
\Comment{\parbox[t]{.5\linewidth}{$\mathcal{A}$ set of weights for the features (models)
	  from $\{f^{i} (x)\}_{i=1}^p$ selected by \bst}}
	  
\State $\mathcal{S} \leftarrow  \emptyset$ 
\Comment{\parbox[t]{.5\linewidth}{$\mathcal{S}$ set of features (models) from $\{f^{i} (x)\}_{i=1}^p$
		selected by \bst}}
\Statex
\State $\beta \leftarrow  0$ \Comment{\parbox[t]{.5\linewidth}{Bias of the ensemble obtained by \bst}}
\Statex
\State $r^{(0)} \leftarrow  y$ \Comment{\parbox[t]{.5\linewidth}{The initial residual in \bst\ is the original target $y$}}
\While{true} 
\State $i^{*,(j)} \leftarrow \arg\min\limits_{\{i: f^{i} (x)\in \mathcal{F}\}}\{l_{i}^{(j)}: [\alpha_{i}^{(j)},\beta_{i}^{(j)},l_{i}^{(j)}]\leftarrow$
\ols$(f^{i} (x),r^{(j-1)},\mathcal{L})\}$ 
\Statex \Comment{\parbox[t]{.5\linewidth}{\bst\ selects the feature with the lowest value of the 
loss function $\mathcal{L}$, performing an \ols\  of each 
feature isolated from the rest.}}  
\Statex
\If {stop\_criterion}      \Comment{\parbox[t]{.5\linewidth}{Stop criterion: \aic, \aicc, \bic, \hqic, \gmdl\ or \icm}} 
%\If {$|\alpha_{i^{*,(j)}}^{(j)}|>|\alpha_{i^{*,(j-1)}}^{(j-1)}|$}      \Comment Stop criterion
\Break
\EndIf
\State $r^{(j)} \leftarrow r^{(j-1)}-(\alpha_{i^{*,(j)}}^{(j)}\cdot f^{i^{*,(j)}} (x)+\beta_{i^{*,(j)}}^{(j)})$  
\Statex \Comment{\parbox[t]{.5\linewidth}{The residual (the target for the next stage) is updated}}
\State $\mathcal{S}  \leftarrow \mathcal{S}\cup\{f^{i^{*,(j)}} (x)\}$  \Comment{\parbox[t]{.5\linewidth}{The selected feature in each stage is added to $\mathcal{S}$ }}
\State $\mathcal{A}  \leftarrow \mathcal{A}\cup\{\alpha_{i^{*,(j)}}^{(j)}\}$  \Comment{\parbox[t]{.5\linewidth}{The weight for the selected feature is added to $\mathcal{A}$}}
\State $\beta  \leftarrow \beta+\beta_{i^{*,(j)}}^{(j)}$  \Comment{\parbox[t]{.5\linewidth}{The bias is updated}}
\EndWhile
\State \Return $\mathcal{S}$, $\mathcal{A}$, $\beta$  \Comment{\parbox[t]{.5\linewidth}{The stop criterion is satisfied and $\mathcal{S}$, $\mathcal{A}$  and $\beta$ are returned}} 
\EndFunction
\end{algorithmic}
\end{algorithm}
\end{figure}
\paragraph{Choosing the feature in each stage, the regressor and the loss function} The criterion applied in order to select a feature in each stage is defined in terms of a loss function $\mathcal{L}$, which will be the same one to be optimized in the HPO process. Hence, the criterion adopted will be the usual one: choosing the feature that produces the lowest value for this loss function \cite{buhlmann2007boosting}. This criterion is expressed as:
\[
f_{i^{*,(j)}}^{(j)}(x)=\arg\min_{f(x)\in \mathcal{F}} \mathcal{L}(r^{(j-1)},\alpha_f^{(j)}\cdot f(x)+\beta_f^{(j)})
\]

Regarding the regressor employed in the process, \ols\ is now adequate, since i) it has no hyperparameters to tune and ii) the regression is performed over just one feature each time, so that, the problems derived from multicollinearity disappear. Consequantly, the loss function $\mathcal{L}$ to minimize will be the typical squared-error $L_2$ loss function \cite{friedman01greedy}.

\paragraph{The algorithm}
Algorithm \ref{alg:bst} displays the pseudocode of the \bst\ procedure. Only one feature is involved in each stage (see the first argument of the call to the \ols\ function in line 7). Hence, just one $\alpha$-coefficient is provided in each stage. In the end, every $\alpha$-coefficient computed in each stage is returned (see lines 13 and 16). The values of the $\beta$ bias computed in all the stages, are added up in order to obtain the final value (see lines 14 and 16). Also, the target varies from one stage to another (see line 5 for the initial target, the second argument of the call to the \ols\ function in line 7, and the target is updated for the next stage in line 11).

Let us now discuss the novel stop criterion and the proposed implicit regularization included in the process.

\subsection{Increasing Coefficient Magnitude as stop criterion}\label{sec:icm}
Concerning the stop criterion, when the selected feature in a given stage, despite being the most highly correlated to the target in said stage, is even poorly correlated to the target, the weight of this feature approaches to zero and therefore the residual of this stage will be close to the that of the previous stage. This fact opened the door to including a heuristic to establish a stop criterion. One may suggest stopping when the loss estimation increases from one stage to another. But this never take place with \ols\ as the regressor and when $L_2$ function is taken as the loss function. When this is the case, the loss estimation is always dismissed from one stage to another, since \ols\ obtains the linear function with precisely the minimum $L_2$ function value. Instead, our proposal consists of establishing a stop criterion in terms of the selected feature coefficients. The existing stop criteria \aic, \aicc, \bic, \hqic\ and \gmdl\ basically depend on the error value, the number of features and the number of examples. But in this context, the number of examples is a constant. This means that only the error value and the number of features have any influence. In the case of \fsr, \pcr\ and \pls, a different feature is added in each iteration. Hence, the error value is the element that conditions when the algorithm stops. Clearly, a feature is highly relevant if the error considerably diminishes. However, if that error is just reduced slightly, these stop criteria may not be able to distinguish whether a feature provides promising information or not, causing an overfitting situation. The same happens with \bst\ and \rbst, but an additional issue affects these ensemble strategies. In fact, both \bst\ and \rbst\ are capable of taking the same feature more than once. Then, if a feature chosen in a certain stage has already been chosen in a previous stage, these stop criteria will always admit this choice because i) since the feature is not a new one the number of features remains constant and, as stated before, ii) the error value always decreases from one stage to the next. Hence, these stop criteria are not useful under this particular situation. In this sense, the fact that our novel stop criterion includes the selected feature's coefficient proves useful when the feature is selected more than once. This is so because the coefficients vary greatly with each selection of the same feature.

Let us now deduce the novel stop criterion. In \ols, the coefficient of the regression $\alpha$ can be expressed in terms of the feature and target standard deviations $\sigma_{f(x)}$ and $\sigma_y$ as \cite{bingham2010regression}
\[
\alpha=R\cdot\sigma_y/\sigma_{f(x)}
\]
where $R^2$ is the correlation coefficient or coefficient of determination. It represent the share of the variation of $y$ that can be explained through the regression model, and it also satisfies $R^2=\sigma_{f(x),y}^2$. Taking into account the ANOVA decomposition, the total variability $SST$ is the sum of the variability associated with the model $SSM$ and the variability of the residuals $SSR$, that is, $SST=SSM+SSR$, where $SST$, $SSM$ and $SSR$ are expressed as follows
\[
SST=\sum_{i=1}^n(y_i-\overline{y})^2\quad  SSM=\sum_{i=1}^n(\hat{y}_i-\overline{y})^2\quad SSR=\sum_{i=1}^n(y_i-\hat{y}_i)^2
\]
where $y_i$ are the actual target values, $\hat{y}_i$ are the predictions, $\overline{y}$ is the target value average, and $n$ is the number of instances.

Then, $SSR/SST$ is the proportion of the variation in the target that is not explained by the regression model. Therefore, $R^2$ can be expressed as
\[
R^2=SSM/SST=1-SSR/SST
\]
Turning back to the expression of $\alpha$ in terms of the correlation coefficient, one can state that
\[
\alpha\cdot \sigma_{f(x)}=R\cdot\sigma_y=\pm\sqrt{\left(1-\frac{SSR}{SST}\right)\cdot\sigma_y^2}
\] 
Since $\sigma_y^2=SST/n$, then
\[
\alpha\cdot \sigma_{f(x)}=\pm\sqrt{\left(1-\frac{SSR}{SST}\right)\cdot\frac{SST}{n}}
\] 
Taking into account that $SST\neq 0$, then
\[
\alpha\cdot \sigma_{f(x)}=\pm\sqrt{\frac{SST-SSR}{n}}
\] 

Let us notice that $SST$ is invariant regardless of the regression model, since $SST$ is the variability contained in the data. Besides, $n$ is also constant. Considering that $SSR$ is positive, the expression $|\alpha|\cdot\sigma_{f(x)}$ ($\sigma_{f(x)}$ is always positive) is maximum when $SSR$ is minimum. Therefore, as \bst\ gets the minimum value for $SSR$,  it also obtains the maximum value for $|\alpha|\cdot\sigma_{f(x)}$. This means that $|\alpha|\cdot\sigma_{f(x)}$ in stage $j-1$ for the selected feature in stage $j-1$ ($f^{i^{*,(j-1)}}(x)$) is greater than or equal to $|\alpha|\cdot\sigma_{f(x)}$ for the rest of the features, including the selected feature in stage $j$ ($f^{i^{*,(j)}}(x)$); otherwise, the selected feature in stage $j-1$ will be a different feature from $f^{i^{*,(j-1)}}(x)$, for instance, the one selected in stage $j$ ($f^{i^{*,(j)}}(x)$). Then,

\[
|\alpha_{i^{*,(j-1)}}^{(j-1)}|\cdot\sigma_{f^{i^{*,(j-1)}}(x)}\geq |\alpha_{i^{*,(j)}}^{(j-1)}|\cdot\sigma_{f^{i^{*,(j)}}(x)}\footnote{Notice that $\alpha_{i^{*,(j)}}^{(j-1)}$ is the coefficient of the regression performed in stage $j-1$ for the feature selected in stage $j$, which may not be the best option in stage $j-1$.}
\]

On the other hand, $|\alpha|\cdot\sigma_{f(x)}$ typically decreases from one stage to the next. This is because the variability of the residuals ($SSR$) always decreases in each stage, since the successive target values contain less information attributable for features as the algorithm progresses. Conversely, an increase of $|\alpha|\cdot\sigma_{f(x)}$ from one stage to another is a sign of a poor $SST$ decrease. If this situation takes place, then $|\alpha|\cdot\sigma_{f(x)}$ in stage $j$ for the selected feature in stage $j$ ($f^{i^{*,(j)}}(x)$) is greater than or equal to $|\alpha|\cdot\sigma_{f(x)}$ in stage $j-1$ for the selected feature in stage $j-1$ ($f^{i^{*,(j-1)}}(x)$), that is
\[
|\alpha_{i^{*,(j)}}^{(j)}|\cdot\sigma_{f^{i^{*,(j)}}(x)}>|\alpha_{i^{*,(j-1)}}^{(j-1)}|\cdot\sigma_{f^{i^{*,(j-1)}}(x)}
\]
Therefore, taking into account the previous inequality leads us to the following one
\[
|\alpha_{i^{*,(j)}}^{(j)}|\cdot\sigma_{f^{i^{*,(j)}}(x)}>|\alpha_{i^{*,(j-1)}}^{(j-1)}|\cdot\sigma_{f^{i^{*,(j-1)}}(x)}\geq |\alpha_{i^{*,(j)}}^{(j-1)}|\cdot\sigma_{f^{i^{*,(j)}}(x)}
\]
Hence,
\[
|\alpha_{i^{*,(j)}}^{(j)}|\cdot\sigma_{f^{i^{*,(j)}}(x)}> |\alpha_{i^{*,(j)}}^{(j-1)}|\cdot\sigma_{f^{i^{*,(j)}}(x)}
\]
Now, and since $\sigma_{f^{i^{*,(j)}}(x)}$ is positive, the following inequality holds:
\[
|\alpha_{i^{*,(j)}}^{(j)}|> |\alpha_{i^{*,(j)}}^{(j-1)}|
\]

This means that the coefficient of the same feature in the current stage is greater than its coefficient in the previous stage. At this point, one can only wonder why this feature takes this greater value in the current stage and has not in the previous stage. Hence, the fact that the influence of a feature increases from one stage to another, and taking into account that $SSR$ always decreases  (sometimes poorly) from one stage to another can be interpreted as a sign that the model overfits the data. Consequently, the proposed stop criterion \icm, aims precisely to prevent this kind of situation. As a result, the algorithm will stop when the following inequality holds
\[
|\alpha_{i^{*,(j)}}^{(j)}|\cdot\sigma_{f^{i^{*,(j)}}(x)}>|\alpha_{i^{*,(j-1)}}^{(j-1)}|\cdot\sigma_{f^{i^{*,(j-1)}}(x)}
\]

Given that the typical deviations $\{\sigma_f{}\}_{f \in \mathcal{F}}$ remain constant during the \bst\ process, \icm\ can be expressed as
\[
\icm_{\{\sigma_{f(x)}\}_{f(x) \in \mathcal{F}}}\left(\alpha_{i^{*,(j)}}^{(j)},\alpha_{i^{*,(j-1)}}^{(j-1)}\right)\equiv\left[|\alpha_{i^{*,(j)}}^{(j)}|\cdot\sigma_{f^{i^{*,(j)}}(x)}>|\alpha_{i^{*,(j-1)}}^{(j-1)}|\cdot\sigma_{f^{i^{*,(j-1)}}(x)}\right]
\]
 
Notice that $\alpha_{i^{*,(j-1)}}^{(j-1)}$ does not exist when $j=1$. In this case, the stop criterion is defined as $\icm_{\{\sigma_{f(x)}\}_{f(x) \in \mathcal{F}}}\left(\alpha_{i^{*,(1)}}^{(1)},\alpha_{i^{*,(0)}}^{(0)}\right)\equiv False$. This means that this stop criterion guarantees the selection of at least one feature.
\newcommand{\nonl}{\renewcommand{\nl}{\let\nl\oldnl}}% Remove line number for one line
 \begin{figure}[p]
      \centering
\begin{algorithm}[H]\small
\caption{\rbst\ for stacking ensemble}
\label{alg:regularizedbst}
\renewcommand{\algorithmicrequire}{\textbf{Input:}}
\renewcommand{\algorithmicensure}{\textbf{Output:}}
\newcommand{\Break}{\State \textbf{break} }
\begin{algorithmic}[1]
\Function{Regularized\_Boosting}{}
\begin{flushleft}
       \textbf{Input:} $\{f^{i} (x)\}_{i=1}^p$ features of the ensemble (model predictions)$, y$ the original target$,\mathcal{L}$ a loss function\\
        \textbf{Output:} $\mathcal{S}$, $\mathcal{A}$, $\beta$
\end{flushleft}
\State $\mathcal{A} \leftarrow  \emptyset$ 
\Comment{\parbox[t]{.5\linewidth}{$\mathcal{A}$ set of weights for the features (models)
	  from $\{f^{i} (x)\}_{i=1}^p$ selected by \rbst}}	  
\State $\mathcal{S} \leftarrow  \emptyset$ 
\Comment{\parbox[t]{.5\linewidth}{$\mathcal{S}$ set of features (models) from $\{f^{i} (x)\}_{i=1}^p$
		selected by \rbst}}
\Statex
\State $\beta \leftarrow  0$ \Comment{\parbox[t]{.5\linewidth}{Bias of the ensemble obtained by \rbst}}
\Statex
\State $r^{(0)} \leftarrow  y$ \Comment{\parbox[t]{.5\linewidth}{The initial residual in \rbst\ is the original target $y$}}
\While{true} 
\State $i^{*,(j)} \leftarrow \arg\min\limits_{\{i: f^{i} (x)\in \mathcal{F}\}}\{l_{i}^{(j)}: [\alpha_{i}^{(j)},\beta_{i}^{(j)},l_{i}^{(j)}]\leftarrow$
\ols$(f^{i} (x),r^{(j-1)},\mathcal{L})\}$ 
\Statex \Comment{\parbox[t]{.5\linewidth}{\bst\ selects the feature with the lowest value of the 
loss function $\mathcal{L}$, performing an \ols\  of each 
feature isolated from the rest.}}  
\Statex
\If {stop\_criterion}      \Comment{\parbox[t]{.5\linewidth}{Stop criterion: \aic, \aicc, \bic, \hqic, \gmdl\ or \icm}} %\If {$|\alpha_{i^{*,(j)}}^{(j)}|>|\alpha_{i^{*,(j-1)}}^{(j-1)}|$}      \Comment Stop criterion
\Statex
\State $\mathcal{A}  \leftarrow \mathcal{A}\setminus\{p_L(j-1)\cdot\alpha_{i^{*,(j-1)}}^{(j-1)}\}\cup\{\alpha_{i^{*,(j-1)}}^{(j-1)}\}$
\Break
\EndIf
\State $r^{(j)} \leftarrow r^{(j-1)}-(p_L(j)\cdot\alpha_{i^{*,(j)}}^{(j)}\cdot f^{i^{*,(j)}} (x)+\beta_{i^{*,(j)}}^{(j)})$ 
\Statex \Comment{\parbox[t]{.5\linewidth}{The residual (the target for the next stage) is updated}}
\State $\mathcal{S}  \leftarrow \mathcal{S}\cup\{f^{i^{*,(j)}} (x)\}$  \Comment{\parbox[t]{.5\linewidth}{The selected feature in each stage is added to $\mathcal{S}$ }}
\State $\mathcal{A}  \leftarrow \mathcal{A}\cup\{p_L(j)\cdot\alpha_{i^{*,(j)}}^{(j)}\}$ \Comment{\parbox[t]{.5\linewidth}{The weight for the selected feature is added to $\mathcal{A}$}}

\State $\beta  \leftarrow \beta+\beta_{i^{*,(j)}}^{(j)}$  \Comment{\parbox[t]{.5\linewidth}{The bias is updated}}
\EndWhile
\State \Return $\mathcal{S}$, $\mathcal{A}$, $\beta$  \Comment{\parbox[t]{.5\linewidth}{The stop criterion is satisfied and $\mathcal{S}$, $\mathcal{A}$  and $\beta$ are returned}} 
\EndFunction
\end{algorithmic}
\end{algorithm}
\end{figure}

\subsection{Implicit regularization}\label{sec:regul}
The \car\ method has been shown to perform well in ensemble HPO. This means that combining several features, even if they are highly correlated can be promising. \bst\ presents a drawback in this respect because it tries to extract the maximum amount of information from each selected feature in each stage while preventing other highly correlated features from bearing any influence on. An implicit regularization is proposed in order to overcome this drawback and make it possible to include correlated features in the ensemble, which may improve the predictive performance. An a priori idea may consists to be weight the coefficient influence using the probability of this feature being relevant in the ensemble, taking into account that all the features selected before have been included in the ensemble. A probability of $1$ for all the features results in the \bst\ approach, since all the selected features have the maximum influence on the ensemble. An alternative stems from the well-known sunrise problem formulated by Laplace, which consists of estimating the probability that the sun will rise tomorrow given that it has previously risen $j-1$ times. This estimation was solved by Laplace himself through his own rule of succession \cite{chung2006elementary}. This probability has been stated as:
\[
p_L(j)=\frac{(j-1)+1}{(j-1)+2}=\frac{j}{j+1}
\]

Initially, the first selected feature will have an influence of $1/2$, allowing in successive stages for other highly correlated features to be selected (including this very same feature). We must remember that each successive target of \bst\ in the following stages are the resulting residuals and hence, the regression coefficients using \ols\ keep decreasing. Taking this into consideration, weighting the coefficients with this probability will avoid having to reduce the influence of those features that are successively selected in more advance stages (even when they are previously selected), since, this probability asymptotically increases to $1$. This implicit regularization is then included when it comes to compute the target (residual) of the next stage. In addition to that, this regularization satisfies being independent from every other stage. This condition is mandatory for \bst\ because \bst\ requires that the regularization procedure may be applied to each selected feature independently of all the features from the remaining stages. This is different from applying the procedure jointly, such as in \gem, which globally adjusts the influence of the features in order to satisfy the constraints of being positive and sum to one. Furthermore, this regularization also satisfies the condition of not including a priori information. This condition is relevant in HPO since there is no a priori available information in this context to be included in the regularization procedure. In this sense, typical regularization procedures either include hyperparameters whose values must be a priori fixed or impose certain conditions to be satisfied  beforehand. Algorithm \ref{alg:regularizedbst} displays \rbst. Particularly, lines 11 and 13 of Algorithm \ref{alg:bst} become, respectively,  line 12 and 14 of Algorithm \ref{alg:regularizedbst}, where $p_L(j)$ weights $\alpha_{i^{*,(j)}}^{(j)}$. Line 9 is included in Algorithm \ref{alg:regularizedbst}. This is because if the stop criterion is satisfied in stage $j$, then the influence of the feature selected in the previous stage $j-1$ will have all the influence. Hence, the correspondent coefficient must go back to its original value.

\section{Experimental settings and multicollinearity analysis}\label{sec:settingsmulticol}
This section covers the experimental settings (see Section \ref{sec:settings}) and also describes and discusses a multicollinearity analysis carried out (see Section \ref{sec:multicollinearity}).

\subsection{Experimental settings}\label{sec:settings}
This section goes through the settings established for the experiments, namely, the datasets, the base-learners with their hyperparameters to be tuned, the methods for performing the ensemble and the evaluation loss description. All code was implemented in Python language using the scikit-learn library\footnote{\url{https://scikit-learn.org/stable/}}.

\paragraph{Datasets}
Datasets coming from the UCI repository\footnote{\label{UCI}\url{https://archive.ics.uci.edu/ml/datasets.php}} were taken for performing the experiments. Table \ref{tab:datasets} displays the number of instances and features of every dataset. The values for both properties vary from one dataset to another, which enables having different scenarios in order to check the behaviour of each approach. Specifically, the number of instances ranges from 100 to 4898 and the number of features varies from 4 to 119. 
\begin{table}[h]
\begin{center}
\caption{Number of instances and features for UCI repository datasets\tablefootnote{Flow and Slump refer to the two outputs of the so-called Slump dataset of the UCI repository for multioutput regression. Notice that both have the same number of instances and features.}}
\begin{tabular}{lrr|lrr}
\hline
 Dataset & Inst. & Feat.&Dataset & Inst. & Feat.\\
\hline
Abalone & 4177 & 11 & Forest & 517 & 13\\
Airfoil Self Noise & 1503 & 6 & Qsar & 908 & 7\\
Auto MPG & 392 & 8 & Servo & 167 & 5\\
Automobile & 158 & 26 & Slump & 103 & 8 \\
Concrete Data & 1030 & 9 &Traffic & 135 & 18 \\
Com. and crime & 1993 & 119  & Red wine quality & 1599 & 12 \\
Fertility & 100 & 10 &White wine quality & 4898 & 12\\
Flow & 103 & 8 & &  & \\
\hline
\end{tabular}
\label{tab:datasets}
\end{center}
\end{table}
\paragraph{Sampling hyperparameter strategies}
Several sampling hyperparameter strategies have been included in the experiments. The approaches chosen were those popular and widespread in the AutoML and CASH frameworks. Two of these strategies are Grid Search (\gs) \cite{claesen2015hyperparameter, sambridge1986novel} and Random Search (\rs) \cite{bergstra2012random}, which are the kind that do not perform a guided search. On the other hand, Bayesian Optimization (\bo) \cite{snoek2012practical}, Particle Swarm Optimization (\pso) \cite{shi1998parameter} and Hyperband (\hb) \cite{li2017hyperband} are part of the methods that carry out a guided search.

\paragraph{Base-learners and their hyperparameters}
Different base-learners were tested in the experiments. Particularly, the approaches taken were Ridge Regression (\ridge)\footnote{\url{https://scikit-learn.org/stable/modules/generated/sklearn.linear_model.Ridge.html}}, Support Vector Regressor (\svr)\footnote{\url{https://scikit-learn.org/stable/modules/generated/sklearn.svm.SVR.html}} and Random Forests Regression (\rfr)\footnote{\url{https://scikit-learn.org/stable/modules/generated/sklearn.ensemble.RandomForestRegressor.html}}. Table \ref{tab:baseline} displays the hyperparameter configuration trials for each sampling strategy. A total of $6\cdot 6=36$ trials for \ridge, $7+7\cdot4=35$ trials for \svr\ and $7\cdot 5=35$ trials for \rfr\ are explored for the \gs\ sampling method. The same number of trials were taken for the \rs\ sampling approach using a uniform distribution on the specified sets. Finally, the same number of iterations was defined for the search over the specified sets of hyperparameter values that were fixed for \bo, \pso\ and \hb.

\begin{table}[h]
\centering
\begin{center}
\caption{Hyperparameter values for the base-learners: \ridge\ (alpha: Regularization strength and solver: Solver used in the computational routines). \svr\ (C: Regularization hyperparameter, kernel: Kernel type used and gamma: RBF kernel parameter).  \rfr\ (min\_samples\_leaf: The minimum fraction of samples at a leaf node and max\_features: The fraction of features in the best split).}
\tabcolsep=0.09cm
\scalebox{0.75}{
\begin{tabular}{lcccc}
    \cline{3-5}
    \multicolumn{2}{l}{} & \multicolumn{3}{c}{Hyperparameter Values}\\
\cline{2-5}
\multicolumn{1}{l}{\multirow{2}{*}{}} & \multicolumn{1}{l}{Hyperparameter} & \multicolumn{1}{c}{\multirow{2}{*}{\gs}} &  \multicolumn{1}{c}{\multirow{2}{*}{\rs}} & \multicolumn{1}{c}{\multirow{2}{*}{\bo, \pso\ and \hb}}\\
&name&&&\\
\hline
\multirow{3}{*}{\ridge} & alpha & $\{0, 10^{[-4:0]}\}$ & $\text{U}([0,1])$&$[0,1]$\\
& \multirow{2}{*}{solver}    & $\{$svd, cholesky,  lsqr,& $\text{U}(\{$svd, cholesky,  lsqr,&$\{$svd, cholesky,  lsqr,\\
&  & sparse\_cg, sag, saga$\}$ &  sparse\_cg, sag, saga$\})$&sparse\_cg, sag, saga$\}$\\
\hline
\multirow{3}{*}{\svr} & C & $\{10^{[-3:3]}\}$ & $10^{\text{U}([-3,3])}$&$[10^{-3}, 10^{3}]$\\
& kernel & $\{$linear, RBF$\}$ &$\text{U}(\{$linear, RBF$\}$)&$\{$linear, RBF$\}$\\
& gamma  &$\{10^{[-3:0]}\}$ &$\text{U}([0.01, 1])$&$[0.01, 1]$\\
\hline
\multirow{2}{*}{\rfr} &min\_samples\_leaf & $\{2^{[-2:-8]}\}$ & $2^{\text{U}([-8,-2])}$&$[2^{-8},2^{-2}]$\\
& max\_features & $\{1, 0.8, 0.6, 0.4, 0.2\}$ &\text{U}([0.2,1])&$[0.2,1]$\\
\hline
\end{tabular}}
\label{tab:baseline}
\end{center}
\end{table}

\paragraph{Ensemble methods}
Several non-hyperparametric stop criteria were checked for Forward Stepwise Regression (\fsr), Principal Component Regression (\pcr), Partial Least Squares (\pls), Boosting (\bst) and Regularized Boosting (\rbst).  Specifically, Akaike Information Criterion (\aic) \cite{akaike1998information}, Akaike Information Criterion corrected (\aicc) \cite{hurvich1989regression}, Bayesian Information Criterion (\bic) \cite{schwartz1997stochastic}, Hannan-Quinn Information Criterion (\hqic) \cite{hannan1979determination} and generalized Minimum Description Length (\gmdl) \cite{hansen2001model} were compared. The best stop criterion among \aic, \aicc, \bic, \hqic\ and \gmdl\ for \fsr, \pcr, \pls, \bst\ and \rbst\ were compared with the proposed stop criterion called Increasing Coefficient Magnitude (\icm), with the Ordinal Least Squares (\ols) and the Generalized Ensemble Method (\gem). Also, a comparison was carried out among the method that just chooses a model (\best), the Basic (\bem) and Generalized (\gem) Ensemble methods \cite{perrone1992networks}, the method that provides weights inversely proportional to the expected error (\inv) \cite{alaa2018autoprognosis} and the \car\ \cite{caruana2004ensemble} method.
%Hence, if we are talking about a 3-cross validation, then, a total of 7950*3= 23850 models are trained to feed the ensemble.

\paragraph{Evaluation score}
The evaluation score was the relative mean squared error computed using a 3-fold cross validation procedure. A higher number of folds would probably provide a lower prediction error. But would also mean training an excessive number of models, which can be computationally heavy and time-consuming. This is especially relevant in this context, since the cross validation is performed along the whole process of Figure \ref{fig:overallschema}. Particularly, in the phase previous to ensemble, a total of 3 base-learners (\ridge, \svr\ and \rfr) were trained. Respectively, a total of 36, 35 and 35 configuration trials (see Table \ref{tab:baseline}) were generated with 5 different sampling strategies (\gs, \rs, \bo, \pso\ and \hb) for 15 different datasets. This leads to $(36+35+35)\cdot 5\cdot 15=7950$ cross validation experiments. In the ensemble phase, there are 5 ensemble strategies different from stacking (\bem, \inv, \gem, \ols\ and \car). In the case of the stacking ensemble method, there are 5 different meta-learners (\fsr, \pcr, \pls, \bst\ and \rbst), each one with 5 stop criteria (\aic, \aicc, \bic, \hqic\ and \gmdl). Additionally, the meta-learners \bst\ and \rbst\ also work with the \icm\ stop criterion, which adds up 2 more ensemble possibilities. In total, there are $5+5\cdot5+2=32$ ensemble strategies. All these ensemble strategies were performed for the 3 base-learners (\ridge, \svr\ and \rfr) and the 5 different sampling strategies (\gs, \rs, \bo,\pso\ and \hb), and 15 datasets, which leads to $3\cdot 5\cdot 15\cdot 32=7200$ cross validation ensembles. Therefore, the 7950 experiments of the phase previous to ensemble and the 7200 ensembles had to be repeated as many times as the number of folds. Hence, and in the interest to compare the approaches rather than to get optimal predictions, a cross validation of just 3 folds was performed in order to reduce the magnitude of the experiments. 

\paragraph{Statistical significant test methods} A Friedman test that rejects the null hypothesis that states that not all learners perform equally \cite{friedman1937use, friedman1940comparison} has been carried out over the evaluation score. The Friedman test is a non-parametric hypothesis test that ranks
all algorithms for each data set separately. If the null-hypothesis (all ranks
are not significantly different) is rejected, the Nemenyi test \cite{nemenyi1962distribution} is adopted as
the post-hoc test. According to the Nemenyi test, the performance of two
algorithms is considered significantly different if the corresponding average
ranks differ by at least the so-called critical difference.

\subsection{Multicollinearity analysis}\label{sec:multicollinearity}
We have argued before how multicollinearity is an issue to avoid, or, at least, mitigate, since it may lead to unstable models under small variations \cite{kiers2007comparison}. Despite the predictive performance may be not affected, it does call into question the significance of highly correlated features. This problem looms in HPO, as commented before, in Section \ref{sec:intro}. This is because certain hyperparameter configurations may produce underfitted models that might give similar prediction values for all the instances, and, in addition, may not induce different models, since the same machine learning system is used, with just little variations in its hyperparameter values.
\begin{figure}[h]
\centering
	\includegraphics[scale=0.9]{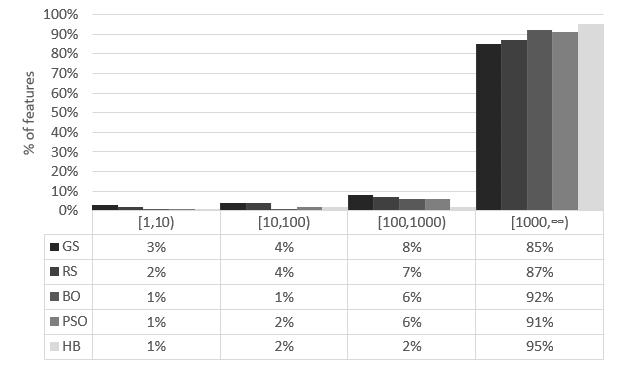}
	\caption{Percentage of features with the specified VIF values among all features taken for the predictions of \ridge, \svr\ and \rfr\ varying all the hyperparameter values explored by \gs, \rs, \bo, \pso\ and \hb.}
	\label{fig:vif}
\end{figure}

Several ways of detecting multicollinearity have been studied \cite{belsley2005regression}, but the Variable Inflation Factor (VIF) has shown to be the most promising and it is the most widely adopted. This is because VIF is based on calculating the linear regression of a single feature directly against the rest of them. The VIF is the inverse of the tolerance. The tolerance is computed as $1-R^2$, where $R^2$ is the coefficient of determination, which measures how well correlated is a certain feature with the remaining ones. $R^2$ indicates the percentage of the variance in a feature that can be attributed to the set of the remaining features. The VIF represents the factor by which the correlations amongst the remaining features contribute to the variance of the feature for which the VIF is computed. This variance is the error in the coefficient estimation. And this error is taken to establish the confidence intervals of the coefficient estimation. Hence, the higher the error, the wider the confidence interval is. Consequently, coefficient estimation becomes unstable and less accurate. The coefficient of determination $R^2$ is computed from the residual sum of squares ($rss$) and the total sum of squares ($tss$) as $R^2=1-\frac{rss}{tss}$. If $R^2$ is equal to $0$, the variance of the remaining independent features cannot be predicted from the independent feature for which the $R^2$ is computed. Therefore, when the VIF is equal to 1, the independent feature for which the $R^2$ is computed is not correlated to the remaining ones, which means multicollinearity does not exist in this regression model. As $R^2$ becomes close to $1$, the independent feature becomes highly correlated with the rest of the features, and multicollinearity tends to infinity. Experience indicates that a VIF greater than 5 or 10 \cite{chatterjee2013regression} indicates multicollinearity \cite{montgomery2006introduccion}. In the case of HPO, the values of VIF drastically exceed these limits, as can be seen in Figure \ref{fig:vif}. Particularly, just $1\%$ or $2\%$ of the features, among all the features taken for the predictions of \ridge\, \svr\ and \rfr\ varying all the hyperparameter values explored by \gs, \rs, \bo, \pso\ and \hb, have a VIF below $10$. Furthermore, between $84\%$ and $95\%$ of the features present values of VIF greater than $1000$.

\begin{table}[p]
 \centering
 \caption{Averaged Friedman ranks for the relative mean squared error over all datasets for \fsr, \pcr, \pls{}, \bst{} and \rbst{}
 using several stop criteria (\aic, \aicc, \bic, \hqic{} and \gmdl), taking into account some base-learners (\ridge{}, \svr{} and \randomforest{}) and several sampling strategies (\gs, \rs, \bo, \pso{} and \hb). The best averaged rank in each row is bolded. \label{tab:stopcriteria}}
\begin{subtable}{}
 \centering
 \tabcolsep=0.06cm
\scalebox{0.5}{
\begin{tabular}{llrrrrr}
\hline
\multicolumn{7}{c}{\fsr} \\ \hline
HPO&MLS&\aic&\aicc&\bic&\hqic&\gmdl\\ \hline
\multirow{3}{*}{\gs}&\ridge&\textbf{3.00}&\textbf{3.00}&\textbf{3.00}&\textbf{3.00}&\textbf{3.00}\\ 
&\svr&\textbf{3.00}&\textbf{3.00}&\textbf{3.00}&\textbf{3.00}&\textbf{3.00}\\ 
&\rfr&\textbf{3.00}&\textbf{3.00}&\textbf{3.00}&\textbf{3.00}&\textbf{3.00}\\ \hline
Mean&\gs&\textbf{3.00}&\textbf{3.00}&\textbf{3.00}&\textbf{3.00}&\textbf{3.00}\\ \hline
\multirow{3}{*}{\rs}&\ridge&\textbf{3.00}&\textbf{3.00}&\textbf{3.00}&\textbf{3.00}&\textbf{3.00}\\ 
&\svr&\textbf{3.00}&\textbf{3.00}&\textbf{3.00}&\textbf{3.00}&\textbf{3.00}\\ 
&\rfr&\textbf{3.00}&\textbf{3.00}&\textbf{3.00}&\textbf{3.00}&\textbf{3.00}\\ \hline
Mean&\rs&\textbf{3.00}&\textbf{3.00}&\textbf{3.00}&\textbf{3.00}&\textbf{3.00}\\ \hline
\multirow{3}{*}{\bo}&\ridge&\textbf{3.00}&\textbf{3.00}&\textbf{3.00}&\textbf{3.00}&\textbf{3.00}\\ 
&\svr&\textbf{3.00}&\textbf{3.00}&\textbf{3.00}&\textbf{3.00}&\textbf{3.00}\\ 
&\rfr&\textbf{3.00}&\textbf{3.00}&\textbf{3.00}&\textbf{3.00}&\textbf{3.00}\\ \hline
Mean&\bo&\textbf{3.00}&\textbf{3.00}&\textbf{3.00}&\textbf{3.00}&\textbf{3.00}\\ \hline
\multirow{3}{*}{\pso}&\ridge&\textbf{3.00}&\textbf{3.00}&\textbf{3.00}&\textbf{3.00}&\textbf{3.00}\\ 
&\svr&\textbf{3.00}&\textbf{3.00}&\textbf{3.00}&\textbf{3.00}&\textbf{3.00}\\ 
&\rfr&\textbf{3.00}&\textbf{3.00}&\textbf{3.00}&\textbf{3.00}&\textbf{3.00}\\ 
Mean&\pso&\textbf{3.00}&\textbf{3.00}&\textbf{3.00}&\textbf{3.00}&\textbf{3.00}\\ \hline
\multirow{3}{*}{\hb}&\ridge&\textbf{3.00}&\textbf{3.00}&\textbf{3.00}&\textbf{3.00}&\textbf{3.00}\\ \hline
&\svr&\textbf{3.00}&\textbf{3.00}&\textbf{3.00}&\textbf{3.00}&\textbf{3.00}\\ 
&\rfr&\textbf{3.00}&\textbf{3.00}&\textbf{3.00}&\textbf{3.00}&\textbf{3.00}\\ 
Mean&\hb&\textbf{3.00}&\textbf{3.00}&\textbf{3.00}&\textbf{3.00}&\textbf{3.00}\\ \hline\hline
Mean&\total&\textbf{3.00}&\textbf{3.00}&\textbf{3.00}&\textbf{3.00}&\textbf{3.00}\\ \hline
\end{tabular}}
\end{subtable}
\begin{subtable}{}
 \centering
 \tabcolsep=0.06cm
\scalebox{0.5}{
 \begin{tabular}{llrrrrr}
 \hline
 \multicolumn{7}{c}{\pcr}\\ \hline
 HPO&MLS&\aic&\aicc&\bic&\hqic&\gmdl\\ \hline
\multirow{3}{*}{\gs}&\ridge&2.80&\textbf{2.47}&2.93&2.77&4.03\\ 
&\svr&3.20&\textbf{2.13}&3.30&2.63&3.73\\ 
&\rfr&2.93&\textbf{2.20}&3.47&2.73&3.67\\ \hline
Mean&\gs&2.98&\textbf{2.27}&3.23&2.71&3.81\\ \hline
\multirow{3}{*}{\rs}&\ridge&3.03&\textbf{2.60}&3.07&\textbf{2.60}&3.70\\ 
&\svr&3.13&\textbf{2.40}&3.13&2.67&3.67\\ 
&\rfr&3.03&\textbf{1.83}&3.67&2.43&4.03\\ \hline
Mean&\rs&3.07&\textbf{2.28}&3.29&2.57&3.80\\ \hline
\multirow{3}{*}{\bo}&\ridge&3.00&\textbf{2.67}&3.10&3.03&3.20\\ 
&\svr&3.07&\textbf{1.90}&3.53&2.60&3.90\\ 
&\rfr&3.00&\textbf{2.07}&3.57&2.87&3.50\\ \hline
Mean&\bo&3.02&\textbf{2.21}&3.40&2.83&3.53\\ \hline
\multirow{3}{*}{\pso}&\ridge&2.97&2.73&3.30&\textbf{2.63}&3.37\\ 
&\svr&3.40&\textbf{2.20}&3.03&2.53&3.83\\ 
&\rfr&3.23&\textbf{2.20}&3.67&2.30&3.60\\ \hline
Mean&\pso&3.20&\textbf{2.38}&3.33&2.49&3.60\\ \hline
\multirow{3}{*}{\hb}&\ridge&3.07&\textbf{2.77}&2.97&2.87&3.33\\ 
&\svr&3.10&\textbf{2.50}&3.30&2.60&3.50\\ 
&\rfr&2.97&\textbf{2.23}&3.50&2.57&3.73\\ \hline
Mean&\hb&3.04&\textbf{2.50}&3.26&2.68&3.52\\ \hline\hline
Mean&\total&3.07&\textbf{2.33}&3.33&2.65&3.62\\ 
\hline
\end{tabular}}
\label{tab:week1}
\end{subtable}
\begin{subtable}{}
 \centering
 \tabcolsep=0.06cm
\scalebox{0.5}{
 \begin{tabular}{llrrrrr}
 \hline
  \multicolumn{7}{c}{\pls}\\ \hline
HPO&MLS&\aic&\aicc&\bic&\hqic&\gmdl\\ \hline
\multirow{3}{*}{\gs}&\ridge&3.00&2.97&\textbf{2.93}&3.00&3.10\\ 
&\svr&2.97&\textbf{2.47}&3.03&3.33&3.20\\ 
&\rfr&3.30&\textbf{2.70}&2.87&2.93&3.20\\ \hline
Mean&\gs&3.09&\textbf{2.71}&2.94&3.09&3.17\\ \hline
\multirow{3}{*}{\rs}&\ridge&2.93&3.27&\textbf{2.77}&3.10&2.93\\ 
&\svr&\textbf{2.83}&2.97&3.13&2.87&3.20\\ 
&\rfr&\textbf{2.67}&2.93&3.23&3.00&3.17\\ \hline
Mean&\rs&\textbf{2.81}&3.06&3.04&2.99&3.10\\ \hline
\multirow{3}{*}{\bo}&\ridge&2.93&\textbf{2.80}&3.13&2.93&3.20\\ 
&\svr&\textbf{2.40}&2.63&3.27&2.83&3.87\\ 
&\rfr&3.23&\textbf{2.67}&2.80&3.03&3.27\\ \hline
Mean&\bo&2.86&\textbf{2.70}&3.07&2.93&3.44\\ \hline
\multirow{3}{*}{\pso}&\ridge&2.97&2.97&\textbf{2.80}&2.97&3.30\\ 
&\svr&3.03&\textbf{2.77}&3.00&3.13&3.07\\ 
&\rfr&2.80&\textbf{2.73}&3.23&2.93&3.30\\ \hline
Mean&\pso&2.93&\textbf{2.82}&3.01&3.01&3.22\\ \hline
\multirow{3}{*}{\hb}&\ridge&\textbf{2.77}&3.10&3.10&\textbf{2.77}&3.27\\ 
&\svr&2.80&\textbf{2.67}&3.27&3.00&3.27\\ 
&\rfr&3.13&2.90&\textbf{2.87}&3.13&2.97\\ \hline
Mean&\hb&2.90&\textbf{2.89}&3.08&2.97&3.17\\ \hline\hline
Mean&\total&2.91&\textbf{2.85}&3.04&2.97&3.23\\    \hline
\end{tabular}}
\label{tab:week1}
\end{subtable}
\begin{subtable}{}
 \centering
 \tabcolsep=0.06cm
\scalebox{0.5}{
 \begin{tabular}{llrrrrr}
\\ \hline
\multicolumn{7}{c}{\bst}\\ \hline
HPO&MLS&\aic&\aicc&\bic&\hqic&\gmdl\\ \hline
\multirow{3}{*}{\gs}&\ridge&\textbf{2.93}&\textbf{2.93}&3.10&\textbf{2.93}&3.10\\ 
&\svr&2.90&2.90&\textbf{2.87}&3.07&3.27\\ 
&\rfr&\textbf{2.93}&\textbf{2.93}&3.10&\textbf{2.93}&3.10\\ \hline
Mean&\gs&\textbf{2.92}&\textbf{2.92}&3.02&2.98&3.16\\ \hline
\multirow{3}{*}{\rs}&\ridge&\textbf{3.00}&\textbf{3.00}&\textbf{3.00}&\textbf{3.00}&\textbf{3.00}\\ 
&\svr&2.73&\textbf{2.43}&3.23&3.10&3.50\\ 
&\rfr&3.13&3.07&\textbf{2.93}&\textbf{2.93}&\textbf{2.93}\\ \hline
Mean&\rs&2.96&\textbf{2.83}&3.06&3.01&3.14\\ \hline
\multirow{3}{*}{\bo}&\ridge&\textbf{3.00}&\textbf{3.00}&\textbf{3.00}&\textbf{3.00}&\textbf{3.00}\\ 
&\svr&2.83&2.83&3.07&\textbf{2.77}&3.50\\ 
&\rfr&\textbf{3.00}&\textbf{3.00}&\textbf{3.00}&\textbf{3.00}&\textbf{3.00}\\ \hline
Mean&\bo&2.94&2.94&3.02&\textbf{2.92}&3.17\\ \hline
\multirow{3}{*}{\pso}&\ridge&\textbf{3.00}&\textbf{3.00}&\textbf{3.00}&\textbf{3.00}&\textbf{3.00}\\ 
&\svr&\textbf{2.93}&3.10&\textbf{2.93}&\textbf{2.93}&3.10\\ 
&\rfr&3.10&3.10&\textbf{2.93}&\textbf{2.93}&\textbf{2.93}\\ \hline
Mean&\pso&3.01&3.07&\textbf{2.96}&\textbf{2.96}&3.01\\ \hline
\multirow{3}{*}{\hb}&\ridge&\textbf{2.93}&\textbf{2.93}&3.10&\textbf{2.93}&3.10\\ 
&\svr&2.83&\textbf{2.73}&3.13&3.10&3.20\\ 
&\rfr&2.97&\textbf{2.87}&3.10&2.97&3.10\\ \hline
Mean&\hb&2.91&\textbf{2.84}&3.11&3.00&3.13\\ \hline\hline
Mean&\total&2.95&\textbf{2.92}&3.04&2.97&3.11\\ 
\hline
\end{tabular}}
 \end{subtable}
\begin{subtable}{}
 \centering
 \tabcolsep=0.06cm
\scalebox{0.5}{
 \begin{tabular}{llrrrrr}
\\ \hline
\multicolumn{7}{c}{\rbst}\\ \hline
HPO&MLS&\aic&\aicc&\bic&\hqic&\gmdl\\ \hline
\multirow{3}{*}{\gs}&\ridge&\textbf{2.87}&3.00&3.00&3.00&3.13\\ 
&\svr&\textbf{2.67}&2.83&3.23&2.97&3.30\\ 
&\rfr&\textbf{2.90}&\textbf{2.90}&3.07&3.07&3.07\\ \hline
Mean&\gs&\textbf{2.81}&2.91&3.10&3.01&3.17\\ \hline
\multirow{3}{*}{\rs}&\ridge&\textbf{2.97}&\textbf{2.97}&\textbf{2.97}&\textbf{2.97}&3.13\\ 
&\svr&3.10&3.10&\textbf{2.77}&2.93&3.10\\ 
&\rfr&\textbf{3.00}&\textbf{3.00}&\textbf{3.00}&\textbf{3.00}&\textbf{3.00}\\ \hline
Mean&\rs&3.02&3.02&\textbf{2.91}&2.97&3.08\\ \hline
\multirow{3}{*}{\bo}&\ridge&\textbf{2.93}&\textbf{2.93}&3.13&\textbf{2.93}&3.07\\ 
&\svr&\textbf{2.83}&3.07&2.90&3.03&3.17\\ 
&\rfr&\textbf{2.87}&\textbf{2.87}&3.03&3.03&3.20\\ \hline
Mean&\bo&\textbf{2.88}&2.96&3.02&3.00&3.14\\ \hline
\multirow{3}{*}{\pso}&\ridge&\textbf{2.93}&\textbf{2.93}&3.07&\textbf{2.93}&3.13\\ 
&\svr&\textbf{2.80}&2.97&3.13&2.97&3.13\\ 
&\rfr&\textbf{3.00}&\textbf{3.00}&\textbf{3.00}&\textbf{3.00}&\textbf{3.00}\\ \hline
Mean&\pso&\textbf{2.91}&2.97&3.07&2.97&3.09\\ \hline
\multirow{3}{*}{\hb}&\ridge&\textbf{2.87}&3.00&3.00&3.00&3.13\\ 
&\svr&2.77&\textbf{2.57}&3.20&3.00&3.47\\ 
&\rfr&\textbf{3.00}&\textbf{3.00}&\textbf{3.00}&\textbf{3.00}&\textbf{3.00}\\ \hline
Mean&\hb&2.88&\textbf{2.86}&3.07&3.00&3.20\\ \hline\hline
Mean&\total&\textbf{2.92}&2.95&3.02&2.99&3.12\\  \hline
\end{tabular}}
\label{tab:week1}
\end{subtable}
\end{table}

\section{Result analysis}\label{sec:results}
This section displays and discusses the performance of the approaches (see Section \ref{sec:performance}) and also analyses the methods in terms of the models involved in the ensemble (see Section \ref{sec:sparsity}).
\subsection{Performance analysis}\label{sec:performance} 
The relative mean squared error, calculated using a 3-fold cross validation procedure, was computed for each dataset, each base-learner (\ridge{}, \svr{} and \randomforest{}), each sampling strategy (\gs, \rs, \bo, \pso{} and \hb) and each ensemble method \fsr, \pcr, \pls{}, \bst{} and \rbst{},
 using several stop criteria (\aic, \aicc, \bic, \hqic{} and \gmdl), each ensemble method \ols, \gem, \bem, \inv, \car\ and the proposed \rbst\ with the novel \icm\ stop criterion. The Friedman ranks of the ensemble methods over each dataset were also computed for all base-learners and for all the sampling strategies. The tables with the relative mean squared error and the corresponding Friedman rank are not displayed in the paper due to lack of space. They are available at https://github.com/laurafernandezdiaz/Ensemble. Instead, Tables  \ref{tab:stopcriteria}, \ref{tab:beststopcriteria} and \ref{tab:bestmethods} show the Friedman ranks averaged over all the datasets. These tables also show the average ranks over the base-learner for each sampling strategy. Finally, the total average ranks over the sampling strategies are displayed at the bottom of the tables. 

Particularly, Table \ref{tab:stopcriteria} shows the averaged Friedman ranks over all datasets for \fsr, \pcr, \pls{}, \bst{} and \rbst{} using several stop criteria (\aic, \aicc, \bic, \hqic{} and \gmdl), taking into account some base-learners (\ridge{}, \svr{} and \randomforest{}) and several sampling strategies (\gs, \rs, \bo, \pso{} and \hb). The results reported in this table enable analyzing the behaviour of the stop criteria when used in different meta-learners. The stop criterion that reports the best performance was selected for each meta-learner in order to be compared to the new proposed stop criterion \icm. As seen, there is not so much difference in performance among the stop criteria. Particularly, \fsr\ reports the same result independently of the stop criterion, base-learner and sampling strategy taken. Analyzing in detail this curious behavior, we found that \fsr\ just takes one feature, the best one, and then all stop criteria are satisfied. This may happen because the best feature (the best prediction) completely explain the target, and consequently, the rest of the predictions (which are all quite similar) do not seem to provide any relevant information. This is not the case in the other methods (see Section \ref{sec:sparsity} about the analysis and study of the features and iterations taken by the methods). In the case of \pcr\ and \pls, the features are combined to obtain the first component, after that, the second component is obtained orthogonally to the first component and so on, which implies that the maximum information of the features is extracted in each step from what had not already been extracted in the previous steps. With \bst\ and \rbst\ the target varies from one stage to the next. In particular, the information explained by the feature selected in each stage is removed from the target in order to obtain the target for the next stage. As a result, the features are taken successively in each stage according to the information contained in the updated target. Both mechanisms guarantee, in a certain sense, that the maximum remaining information can be collected until certain stage, when the stop criterion is satisfied. The ranks of the different stop criteria for \pcr, \pls, \bst\ and \rbst\ are quite similar. In fact, our own stop criterion behaves quite similarly. \aic\ is the point of departure and the rest just add some correction factors to consider the balance between instances and features or whether the number of instances is exceeded. In any case, \aicc\ seems to provides the best results for \pcr\ and \pls. As to \bst\ and \rbst, both \aic\ and \aicc\ slightly outperform the rest. \gmdl\ is the worst, followed by \bic, so penalizing the number of instances does not seem to be a good practice. There are significant differences, up to $90\%$ and $95\%$, \footnote{The respective critical differences for confidence levels of $90\%$ and $95\%$ are $0.36$ and $0.40$} between \aicc\ and \aic, \bic\ and \hqic, but only in the case of \pcr.

\begin{table}[h]
\centering
\caption{Averaged Friedman ranks for the relative mean squared error over all datasets for \ols{}, \gem{} and the best stop criteria among  \aic, \aicc, \bic, \hqic{} and \gmdl{}, for \fsr, \pcr, \pls{}, \bst{} and \rbst{}, as well as the novel stop criterion \icm\ for \bst\ and \rbst, taking into account some base-learners (\ridge{}, \svr{} and \randomforest{}) and several sampling strategies (\gs, \rs, \bo, \pso{} and \hb). The best averaged rank in each row is bolded. \label{tab:beststopcriteria}}
 \tabcolsep=0.3cm
\scalebox{0.65}{
\begin{tabular}{llccccccccc}
\hline
%HPO&MLS&\ols&\gem&\fsr(*)&\pcr(\aicc)&\pls(\aicc)&\bst(\aicc)&\rbst(\aic)&\bst(\icm)&\rbst(\icm)\\ 
\multirow{2}{*}{HPO}&\multirow{2}{*}{MLS}&\multirow{2}{*}{\ols}&\multirow{2}{*}{\gem}&\fsr&\pcr&\pls&\bst&\rbst&\bst&\rbst\\ 
&&&&(*)&(\aicc)& (\aicc)&(\aicc)&(\aic) &(\icm)&(\icm)\\
\hline
\multirow{3}{*}{\gs}&\ridge&8.93&5.53&4.13&7.27&4.53&3.77&4.03&3.73&\textbf{3.07}\\ 
&\svr&8.13&4.60&4.87&7.67&5.00&4.63&4.23&3.33&\textbf{2.53}\\ 
&\rfr&6.27&2.93&5.10&8.00&5.80&5.10&4.93&4.20&\textbf{2.67}\\ \hline
Mean&\gs&7.78&4.36&4.70&7.64&5.11&4.50&4.40&3.76&\textbf{2.76}\\ \hline
\multirow{3}{*}{\rs}&\ridge&8.70&5.03&4.27&6.97&4.57&3.97&4.17&3.90&\textbf{3.43}\\ 
&\svr&8.10&5.17&4.53&7.83&4.17&\textbf{3.40}&4.47&3.57&3.77\\ 
&\rfr&6.53&\textbf{3.53}&4.63&7.80&5.47&5.13&4.63&3.67&3.60\\ \hline
Mean&\rs&7.78&4.58&4.48&7.53&4.73&4.17&4.42&3.71&\textbf{3.60}\\ \hline
\multirow{3}{*}{\bo}&\ridge&9.00&6.07&4.10&6.33&4.33&4.10&3.93&\textbf{3.20}&3.93\\ 
&\svr&7.87&4.13&4.77&7.40&5.40&3.97&4.20&4.00&\textbf{3.27}\\ 
&\rfr&5.53&3.47&5.20&8.33&5.20&5.20&4.87&4.33&\textbf{2.87}\\ \hline
Mean&\bo&7.47&4.56&4.69&7.36&4.98&4.42&4.33&3.84&\textbf{3.36}\\ \hline
\multirow{3}{*}{\pso}&\ridge&9.00&6.33&3.63&6.60&5.20&3.63&3.40&3.87&\textbf{3.33}\\ 
&\svr&7.93&4.13&4.73&8.27&5.33&4.73&4.47&2.87&\textbf{2.53}\\ 
&\rfr&7.60&4.73&4.23&8.27&6.33&4.33&4.23&\textbf{2.60}&2.67\\ \hline
Mean&\pso&8.18&5.07&4.20&7.71&5.62&4.23&4.03&3.11&\textbf{2.84}\\ \hline
\multirow{3}{*}{\hb}&\ridge&8.93&5.67&3.93&7.47&4.47&3.63&3.83&3.67&\textbf{3.40}\\ 
&\svr&7.87&5.47&4.57&7.67&5.00&4.27&3.97&3.87&\textbf{2.33}\\ 
&\rfr&5.47&3.67&5.10&8.00&6.33&4.87&5.17&3.47&\textbf{2.93}\\ \hline
Mean&\hb&7.42&4.93&4.53&7.71&5.27&4.26&4.32&3.67&\textbf{2.89}\\ \hline\hline
Mean&\total&7.60&4.64&4.52&7.61&5.20&4.33&4.33&3.63&\textbf{3.13}\\ \hline
\end{tabular}}
\end{table}

Table \ref{tab:beststopcriteria} displays the averaged Friedman ranks over all datasets for \ols, \gem\footnote{\gem\ is included here for being a version of \ols\ with the constraint of the weights to be positive and sum to one.} and for the best stop criteria (according to the results of Table \ref{tab:stopcriteria}) among  \aic, \aicc, \bic, \hqic{} and \gmdl{} for \fsr, \pcr, \pls{}, \bst{} and \rbst{}, as well as the novel stop criterion \icm\ for \bst\ and \rbst, taking into account some base-learners (\ridge{}, \svr{} and \randomforest{}) and several sampling strategies (\gs, \rs, \bo, \pso{} and \hb). In this table, one can observe that both \bst\ and \rbst\  outperform the rest of the methods, for the best stop criterion among \aic, \aicc, \bic, \hqic{} and \gmdl{} and also for the novel stop criterion \icm\ (see the last row of the last four columns of Table \ref{tab:beststopcriteria}). Indeed, all of them, \bst(\aicc), \rbst(\aic), \bst(\icm) and \rbst(\icm), present significant differences at the confidence levels of $90\%$ and even $95\%$\footnote{The respective critical differences for confidence levels of $90\%$ and $95\%$ are $0.73$ and $0.80$.} when compared to \ols, \pcr\ and \pls. Besides, using \icm\ as stop criterion makes \bst\ and \rbst\ significantly better at these confidence levels than \gem\ and \fsr, in addition to \ols, \pcr\ and \pls. The poor performance of \ols\ may be the result of multicollinearity, which is highly present in HPO. This drawback is corrected by \gem\ regularizing in \ols\ by constraining of the weights to be positive and sum to one. In the case of \fsr, it just happens that regardless of the stop criteria only one feature is selected (see the comments of Table \ref{tab:stopcriteria}), hence multicollinearity disappears. In fact, \gem\ and \fsr\ perform considerably better than \ols. The difference in performance between \pcr\ and \pls\ is caused by the former not taking into account the target in order to build the components, whereas the latter does. These results confirm two well-known conclusions, namely, i) the target contains critical information and ii) regularization helps to alleviate the problems derived from multicollinearity. In this sense, \bst\ and \rbst\ (more so, because \rbst\ includes a regularization procedure) succeed because they squeeze the information contained in the target. In order to achieve this, each stage the information explained by the selected feature is removed from the target, and the remaining information is left to be explained by the features selected in the following stages. Comparing \bst\ and \rbst, \rbst\ stands out as the best option because of the regularization element added to \bst. Hence, smoothing the influence of the features selected in the first stages in order to allow subsequent features to take part of the ensemble clearly improves the overall performance of the ensemble. In addition, the novel stop criterion \icm\ improves both \bst\ and \rbst. The results state that \icm\ is a robust criterion, since it is based on the coefficient from the feature selected  in each stage, which enables it to discern promising choices of features, both when a feature has already been selected or if a feature is selected for the first time. In any case, it seems that \rbst\ benefits from \icm\ to a higher degree than \bst. In fact, \rbst(\icm) works significantly better that \bst (\aicc) and \rbst(\aic), whereas \bst(\icm) does not. Hence, the synergy between both the implicit regularization and the \icm\ helps to improve the predictive performance of the ensemble. Both improvements are independent from each other, which allows applying both of them simultaneously and combining the benefits provided by them.

\begin{table}
\centering
\caption{Averaged  Friedman ranks for the relative mean squared error over all datasets for \best, \bem, \inv, \gem, \car{} and \rbst(\icm), taking into account some base-learners (\ridge{}, \svr{} and \randomforest{}) and several sampling strategies (\gs, \rs, \bo, \pso{} and \hb). The best averaged rank in each row is bolded. \label{tab:bestmethods}}
\tabcolsep=0.3cm
\scalebox{0.65}{
\begin{tabular}{llcccccc}
\hline
%HPO&MLS&\best&\bem&\inv&\gem&\car&\rbst(\icm)\\ \hline
\multirow{2}{*}{HPO}&\multirow{2}{*}{MLS}&\multirow{2}{*}{\best}&\multirow{2}{*}{\bem}&\multirow{2}{*}{\inv}&\multirow{2}{*}{\gem}&\multirow{2}{*}{\car}&\rbst\\ 
&&&&&&&(\icm)\\
\hline
\multirow{3}{*}{\gs}&\ridge&2.97&4.93&4.47&3.67&2.97&\textbf{2.00}\\ 
&\svr&3.40&5.07&4.27&3.47&2.60&\textbf{2.20}\\ 
&\rfr&3.80&4.77&3.87&2.90&\textbf{2.67}&3.00\\ \hline
Mean&\gs&3.39&4.92&4.20&3.34&2.74&\textbf{2.40}\\ \hline
\multirow{3}{*}{\rs}&\ridge&3.20&4.87&4.53&3.37&2.67&\textbf{2.37}\\ 
&\svr&3.53&4.97&3.67&3.87&2.80&\textbf{2.17}\\ 
&\rfr&3.33&4.93&3.93&3.00&\textbf{2.90}&\textbf{2.90}\\ \hline
Mean&\rs&3.36&4.92&4.04&3.41&2.79&\textbf{2.48}\\ \hline
\multirow{3}{*}{\bo}&\ridge&3.10&4.40&4.00&3.67&3.37&\textbf{2.47}\\ 
&\svr&3.60&5.13&4.40&\textbf{2.47}&2.93&\textbf{2.47}\\ 
&\rfr&3.87&4.87&4.07&2.93&2.73&\textbf{2.53}\\ \hline
Mean&\bo&3.52&4.80&4.16&3.02&3.01&\textbf{2.49}\\ \hline
\multirow{3}{*}{\pso}&\ridge&5.20&4.60&3.87&2.93&2.47&\textbf{1.93}\\ 
&\svr&5.93&4.93&3.93&2.53&2.13&\textbf{1.53}\\ 
&\rfr&6.00&4.53&3.87&2.73&2.07&\textbf{1.80}\\ \hline
Mean&\pso&5.71&4.69&3.89&2.73&2.22&\textbf{1.76}\\ \hline
\multirow{3}{*}{\hb}&\ridge&3.40&4.80&4.47&3.33&2.93&\textbf{2.07}\\ 
&\svr&3.67&5.30&3.47&3.90&2.53&\textbf{2.13}\\ 
&\rfr&3.93&4.80&3.73&3.20&\textbf{2.27}&3.07\\ \hline
Mean&\hb&3.67&4.97&3.89&3.48&2.58&\textbf{2.42}\\ \hline\hline
Mean&\total&4.04&4.84&3.98&3.14&2.65&\textbf{2.34}\\ 
\hline
\end{tabular}}
\end{table}

Finally, Table \ref{tab:bestmethods} presents the averaged Friedman ranks over all datasets for \best\ (not performing ensemble), \bem, \inv, \gem\footnote{\gem\ is included here for its performance and as an improvement of \bem.}, \car\ and \rbst\ with the novel stop criterion \icm\ (the best result in Table \ref{tab:beststopcriteria}), again taking into account the same base-learners (\ridge, \svr\ and \rfr) and the same sampling strategies (\gs, \rs, \bo, \pso\ and \hb). Both the \car\ method and \rbst (\icm) clearly outperform \best, \bem, \inv\ and \gem. Besides, the differences at s of $90\%$ and $95\%$\footnote{The respective critical differences for confidence levels for $90\%$ and $95\%$ are $0.45$ and $0.50$.} are significant. \rbst(\icm) performs better than \car\ in almost all cases, but there are no significant differences between them. \best, \bem\ and \inv\ show the worst results. These are very simple ensemble methods that do not include a learning procedure in the ensemble and they select each feature only once. \gem\ performs slightly better, it includes a learning procedure, which adds a higher generalization power. Particularly, it performs an \ols\ with constraints as a regularization procedure, but if a feature is taken, it is taken only once, such as in \best, \bem\ and \inv. Unlike these methods, both \car\ and \rbst\ (as well as \bst) are able to take the same feature more than once, and then fully exploit the information contained in it. Nevertheless, \rbst, like \gem, includes a learning procedure which may add more generalization power and a regularization procedure.

\begin{table}[h]
\centering
\caption{Time (in seconds) of execution for the training of all the 36 (for \ridge) or 35 (for \svr\ and \rfr) models (the sum regarding all datasets) for each base-learner and each sampling strategy}
\label{tab:computationaltimemodels}
\tabcolsep=0.09cm
\begin{tabular}{lrrr|r}\\ \hline
&	\ridge & \svr & \rfr & total\\
\hline
\gs	&29904&	779056&	30725&	839685\\
\rs	&44275&	799202&	43934&	887411\\
\bo	&24230&	770431&	21950&	816611\\
\pso	&25178&	766460&	22795&	814433\\
\hb	&62496&	812348&	64472&	939316\\
\hline
total	&186082&	3927498&	183876&	4297456\\
\hline
\end{tabular}
\end{table}

\begin{table}[h]
\centering
\caption{Time (in seconds) of execution for the ensemble approaches (the sum regarding all base-learners, all datasets and all sampling strategies)}
\label{tab:computationaltimeensemble}
\tabcolsep=0.09cm
\begin{subtable}{}
 \centering
 \tabcolsep=0.25cm
\begin{tabular}{llrrrrrr}\\ \hline
&&\multicolumn{1}{c}{\best} & \multicolumn{1}{c}{\bem} & \multicolumn{1}{c}{\inv} & \multicolumn{1}{c}{\ols} & \multicolumn{1}{c}{\gem} & \multicolumn{1}{c}{\car} \\
\hline
&& 2.25          & 4.50         & 155.25       & 254.25       & 4950.00      & 4050.00               
\end{tabular}
\end{subtable}
\\
\begin{subtable}{}
\centering
 \tabcolsep=0.25cm
\begin{tabular}{lrrrrr} 
 \hline
 \hline
              & \multicolumn{1}{c}{\fsr} & \multicolumn{1}{c}{\pcr} & \multicolumn{1}{c}{\pls} & \multicolumn{1}{c}{\bst} & \multicolumn{1}{c}{\rbst} \\
\hline
\aic  & 207.00       & 560.25       & 2400.75      & 150.75       & 576.00        \\
\aicc & 405.00       & 828.00       & 117.00       & 310.50       & 371.25        \\
\bic & 2517.75      & 175.50       & 274.50       & 171.00       & 193.50        \\
\hqic & 279.00       & 594.00       & 450.00       & 159.75       & 542.25        \\
\gmdl & 623.25       & 801.00       & 423.00       & 162.00       & 378.00        \\
\icm  & \multicolumn{1}{c}{-}            & \multicolumn{1}{c}{-}              & \multicolumn{1}{c}{-}              & 571.50       & 492.75      \\ \hline 
\end{tabular}
\end{subtable}
\end{table}

\subsection{Computational time analysis}
This section deals with a comparison of the ensemble strategies in regard to the computational time. Table \ref{tab:computationaltimemodels} shows the computational time (in seconds) spent on training the models that would feed the ensemble for all datasets and all the configuration trials generated by each sampling strategy (\gs, \rs, \bo, \pso\ and \hb) and for each base-learner (\ridge, \svr\ and \rfr). It also shows the total time spent by each sampling strategy and each base-learner. As it shows, the total computational time spent on training all the models is almost $50$ days (4297456 seconds). Each sampling strategy spent similar computational time. However, the base-learner \svr\  spent considerably much more computational time that \ridge\ and \rfr\ (more than $90\%$ of the computational time). We are reminded that the number of configuration trials for the base-learner are almost equal ($36$ for \ridge\ and $35$ for \svr\ and \rfr). Table \ref{tab:computationaltimeensemble} displays the computational time in seconds for the different ensemble approaches. The time is computed adding up the time spent by all base-learners, all datasets and all sampling strategies. Obviously, the least costly approaches are those that do not include a learning procedure in the ensemble, that is, \best, \bem\ and \inv\ (note that the \best\ method performs no ensemble). Conversely, the most costly approaches are \gem\ and \car. Comparing the computational time reported in Table \ref{tab:computationaltimemodels} with the computational time reported in Table \ref{tab:computationaltimeensemble}, the computational time spent by the ensemble approaches is considerably lower than the computational time spent on training the models for all the configuration trials. This allows concluding there is a great benefit in performance when carrying out an ensemble procedure, in comparison with the small loss in computational time, since the models that feed the ensemble must be trained in any case.

\begin{table}
\caption{Averaged (over all datatasets and all base-learners) number of different features (models) and number of features with replacement (iterations) for the \car\ method, and for \bst\ and \rbst\ taking all the stop criteria.}\label{tab:sparsity}
\centering
 \tabcolsep=0.08cm
\scalebox{0.8}{
% \usepackage{multirow}
\begin{tabular}{lr|rrrrrr|rrrrrr}
\cline{2-14}
&\multicolumn{13}{c}{Different features}\\
\cline{2-14}
                   & \multicolumn{1}{c|}{\multirow{2}{*}{\car}} & \multicolumn{6}{c|}{\bst}                                                                                                                             & \multicolumn{6}{c}{\rbst}                                                                                                                            \\ 
                   \cline{3-14} 
              & \multicolumn{1}{c|}{}                                &  \multicolumn{1}{c}{\aic} & \multicolumn{1}{c}{\aicc} & \multicolumn{1}{c}{\bic} & \multicolumn{1}{c}{\hqic} & \multicolumn{1}{c}{\gmdl} & \multicolumn{1}{c|}{\icm} & \multicolumn{1}{c}{\aic} & \multicolumn{1}{c}{\aicc} & \multicolumn{1}{c}{\bic} & \multicolumn{1}{c}{\hqic} & \multicolumn{1}{c}{\gmdl} & \multicolumn{1}{c}{\icm}  \\ \hline
\multicolumn{1}{l}{\gs}            & 11.20                                                 & 1.17                    & 1.17                     & 1.06                    & 1.14                     & 1.00                     & 3.10                    & 1.19                    & 1.17                     & 1.09                    & 1.15                     & 1.00                     & 3.48                    \\
\multicolumn{1}{l}{\rs}             & 11.65                                                 & 1.17                    & 1.14                     & 1.05                    & 1.09                     & 1.00                     & 3.19                    & 1.09                    & 1.09                     & 1.07                    & 1.07                     & 1.00                     & 3.46                    \\
\multicolumn{1}{l}{\bo}              & 9.54                                                  & 1.20                    & 1.20                     & 1.12                    & 1.17                     & 1.03                     & 2.96                    & 1.18                    & 1.15                     & 1.09                    & 1.12                     & 1.03                     & 3.28                    \\
\multicolumn{1}{l}{\pso}            & 9.33                                                  & 1.22                    & 1.21                     & 1.19                    & 1.19                     & 1.17                     & 3.93                    & 1.25                    & 1.24                     & 1.20                    & 1.24                     & 1.17                     & 3.57                    \\
\multicolumn{1}{l}{\hb}            & 11.95                                                 & 1.17                    & 1.16                     & 1.06                    & 1.15                     & 1.00                     & 3.19                    & 1.21                    & 1.17                     & 1.07                    & 1.12                     & 1.00                     & 3.78                    \\ \hline
\multicolumn{1}{l}{\textbf{Mean}}  & \textbf{10.74}                                      & \textbf{1.19}        & \textbf{1.18}            & \textbf{1.10}          & \textbf{1.15}        & \textbf{1.04}          & \textbf{3.27}        & \textbf{1.18}         & \textbf{1.16}          & \textbf{1.10}          &\textbf{1.14}           & \textbf{1.04}           & \textbf{3.51}   \\  
\hline       
%\end{tabular}}
%\end{table}
%\begin{table}
 %\tabcolsep=0.08cm
%\scalebox{0.9}{
%\begin{tabular}{l|r|rrrrrr|rrrrrr|}
&\multicolumn{13}{c}{Features with replacement}\\
 \cline{2-14}
                   & \multicolumn{1}{c|}{\multirow{2}{*}{\car}} & \multicolumn{6}{c|}{\bst}                                                                                                                             & \multicolumn{6}{c}{\rbst}                                                                                                                            \\ \cline{3-14} 
              & \multicolumn{1}{c|}{}                                &  \multicolumn{1}{c}{\aic} & \multicolumn{1}{c}{\aicc} & \multicolumn{1}{c}{\bic} & \multicolumn{1}{c}{\hqic} & \multicolumn{1}{c}{\gmdl} & \multicolumn{1}{c|}{\icm} & \multicolumn{1}{c}{\aic} & \multicolumn{1}{c}{\aicc} & \multicolumn{1}{c}{\bic} & \multicolumn{1}{c}{\hqic} & \multicolumn{1}{c}{\gmdl} & \multicolumn{1}{c}{\icm}  \\ \hline
\multicolumn{1}{l}{\gs}         & 25.80 &  1.17 &   1.17 &  1.06 &   1.13 &     1.00 &    285.29 &      1.18 &      1.16 &      1.08 &      1.12 &        1.00 &       5.69                   \\

\multicolumn{1}{l}{\rs}        &     24.97 &   1.10 &   1.14 &  1.03 &    1.06 &     1.00 &    302.65 &   1.57 &   1.57 &  1.55 &   1.56 &   1.51 &   5.72      \\
\multicolumn{1}{l}{\bo}     &      23.46 &  1.11 &   1.11 &  1.05 &   1.09 &     1.00 &    423.55 &      1.14 &       1.11 &      1.06 &      1.09 &        1.00 &      5.57                 \\
\multicolumn{1}{l}{\pso}   &   19.51 &  1.01 &   1.02 &  1.01 &   1.01 &     1.00 &    352.93 &      1.05 &      1.04 &      1.02 &      1.04 &        1.00 &      5.43                    \\
\multicolumn{1}{l}{\hb}    &   26.75 &  1.16 &   1.21 &  1.07 &   1.15 &     1.00 &    303.32 &      1.17 &      1.13 &      1.07 &       1.10 &        1.00 &      6.15               \\ \hline   
\multicolumn{1}{l}{\textbf{Mean}} & \textbf{24.10}                                        & \textbf{1.11}           & \textbf{1.13}            & \textbf{1.05}           & \textbf{1.09}            & \textbf{1.00}            & \textbf{333.54}         & \textbf{1.22}           & \textbf{1.20}            & \textbf{1.16}           & \textbf{1.18}            & \textbf{1.10}            & \textbf{5.71}    \\  \hline           
\end{tabular}}
\end{table}

\subsection{Analysis of the models involved in the ensemble}\label{sec:sparsity}
An analysis of the number of models involved in the ensemble was performed. Table \ref{tab:sparsity} displays the averaged (over all datatasets and all base-learners) number of different features (models) taken by the \car\ method, and by \bst\ and \rbst\ for all the stop criteria studied (see the top of Table \ref{tab:sparsity}). It also shows the averaged (again over all datatasets and all base-learners) number of features with replacement taken, which is in fact the number of iterations carried out by the methods (see the bottom of Table \ref{tab:sparsity}). 

Two conclusions can be drawn for the analysis of the different features considered in the ensemble, and: i) the \car\ method is by far the approach that takes the highest number of different features (around 10) and ii) the common stop criteria just select an average of one feature, which is very similar to using the \best\ method. Hence, these stop criteria do not allow neither \bst\ nor \rbst\  to fully exploit the information contained in the models. In this sense, the novel stop criterion \icm\ makes \bst\ and \rbst\ behave conservatively as to how many different features are to be taken in the ensemble compared to the \car\ method, but not as restrictive as with the typical stop criteria.

Regarding the number of features considered in the ensemble, and taking into account the replacement procedure, it is quite noticeable the number of iterations that \bst\ with \icm\ stop criterion spends (in the hundreds) to end up taking only about $3$ different features. However, the number of iterations drastically falls for \rbst\ with \icm. In fact, \rbst\ with \icm\ hardly performs replacement when compared to the \car\ method.

\section{Conclusions and future work}\label{sec:conclusions}
This paper proposes an improved boosting approach as a meta-learner in HPO stacking ensemble, which may be included in an Automated Machine Learning (AutoML) system and which gets better predictive performance. In particular, an implicit regularization would be included in the classical boosting (\bst) method, leading to the method called Regularized Boosting (\rbst). Besides, a novel non-hyperparametric stop criterion for both \bst\ and \rbst\ methods called Increasing Coefficient Magnitude (\icm) is also proposed. Both \rbst\ and \icm\ are specifically designed for HPO. The result is a new meta-learner for stacking ensemble that is shown to be superior to other possible non-hyperparametric (with an adequate non-hyperparametric stop criterion) meta-learners, such as Forward Search (\fsr), Principal Component Regression (\pcr) or Partial Least Squares (\pls). Unlike these methods, \rbst\ with \icm\ is built on the basis of \bst, since our proposal works under the hypothesis that \bst\ is a promising regressor in HPO stacking ensemble. The reason for this is that it performs a regression with just one feature in each stage and uses the successive residuals as targets. These are promising properties, since they allow the use of least squares (which is free of hyperparameter tuning) without being affected by the problems derived from multicollinearity. \icm\ has shown to be a robust criterion, since it is based on the coefficient of the selected feature in each stage rather than only on the error value, number of features and instances, which is what other state-of-the-art stop criteria are typically based on. In fact, \icm\ is able to discern special situations, which other stop criteria are not. The power of \rbst\ lies in fully exploiting the information contained in the target. Specifically, it smoothes the weight of the features selected in the first stages in order to provide other features with an opportunity to supply further information.

It is worth noting that AutoML systems hardly include ensemble in their frameworks. Only some of them do it. The ensemble strategies they typically include are the simple (weighted or not) average or, as in the \car\ method, which computes an average with replacement and has been  widely used among researchers. Additionally, some AutoML systems include stacking ensemble, where the interest lies in the included learning process, which might provide the ensemble with higher generalization power. The main drawback of stacking ensemble is the choice of an adequate meta-learner, for which there is a lack of advice in the literature in general, and  in the AutoML frameworks in particular. Besides, adequate meta-learners may include real-value hyperparameters that need to be tuned in order to avoid the problems derived from multicollinearity. In this respect, the contribution of this paper is not limited to proposing \rbst\ with \icm; in addition, it begins by performing an exhaustive study of possible non-hyperparametric (with an adequate non-hyperparametric stop criterion) meta-learners, such as \fsr, \pcr, \pls\ and even the original \bst, a study that, to the best of our knowledge, has not been carried out in HPO so far. In fact, this study has helped to lay down the foundations for developing the novel approach \rbst\ with the novel stop criterion \icm. In this respect, all those methods feature of the generalization power of a learning procedure and are non-hyperparametric with an adequate non-hyperparametric stop criterion. Moreover, \pcr, \pls, \bst\ and \rbst\ are able to cope with the problems derived from multicollinearity. However, in the case of HPO, \bst\ and \rbst\  provide better performance than \pcr\ and \pls. Furthermore,  \rbst, especially using \icm\ as stop criterion, exhibits superiority, even over \bst, and also with regard to other state-of-the-art ensemble procedures typically included in AutoML frameworks. A computational time study was carried out, concluding that performing an ensemble process is worth, since the additional time spent on the ensemble is considerably lower than the time spent on training the models in order to feed the ensemble, which must be taken into account since these models must be trained in any case.

As future work,  it would be interesting to include the ensemble procedure inside the guided search performed by the sampling strategies such as \bo, \pso\ or \hb. Hence, the next hyperparameter configuration trial in the sampling strategy would be chosen according to the best ensemble from the predictions provided by the models induced using the previous configuration trials. Another proposal for a future line of work would be to provide a non-linear ensemble strategy. Finally, this approach could be extended onto data under distribution changes, such as covariate shift.

\section*{Acknowledgments}\label{sec:Acknowledgments}
This research has been partially supported by the Spanish Ministerio de Ciencia e Innovación through the grant PID2019-110742RB-I00.

\bibliographystyle{elsarticle-num}

\bibliography{mybib}

\end{document}